\documentclass[review,12pt,authoryear]{elsarticle}

\usepackage{amssymb}
\usepackage{amsmath}
\usepackage{multirow}

\journal{Medical Image Analysis}

\begin{document}

\begin{frontmatter}



\title{3DFETUS: Deep Learning-Based Standardization of Facial Planes in 3D Ultrasound} 


\author[1]{Antonia Alomar\corref{cor1}} 
\cortext[cor1]{Corresponding author}
\ead{antonia.alomar@upf.edu}
\author[2]{Ricardo Rubio}

\author[3]{Gerard Albaiges}
\author[4]{Laura Salort-Benejam}
\author[3]{Julia Caminal} 
\author[2]{Maria Prat}
\author[2]{Carolina Rueda}
\author[3]{Berta Cortes}
\author[1]{Gemma Piella}
\author[1]{Federico Sukno}
 
\affiliation[1]{organization={Department of Engineering, Universitat Pompeu Fabra},
            addressline={122-140 Tànger}, 
            city={Barcelona},
            postcode={08018}, 
            country={Spain}}
\affiliation[2]{organization={Department of Obstetrics and Gynecology, Hospital del Mar},
            addressline={25-29 Passeig Marítim}, 
            city={Barcelona},
            postcode={08003}, 
            country={Spain}}  
\affiliation[4]{organization={Department of Robotics, Universitat Politecnica de Catalunya},
            addressline={Carrer de Jordi Girona, 1-3}, 
            city={Barcelona},
            postcode={08034}, 
            country={Spain}}
\affiliation[3]{organization={Gynecology and Reproductive Medicine, University Hospital Quirón Dexeus},
            addressline={Carrer de Sabino Arana, 19}, 
            city={Barcelona},
            postcode={08028}, 
            country={Spain}}
    
\begin{abstract}

The automatic localization and standardization of anatomical planes in 3D medical imaging remains a challenging problem due to variability in object pose, appearance, and image quality. In 3D ultrasound, these challenges are exacerbated by speckle noise and limited contrast, particularly in fetal imaging.

To address these challenges in the context of facial assessment, we present: 1) GT++, a robust algorithm that estimates standard facial planes from 3D US volumes using annotated anatomical landmarks; and 2) 3DFETUS, a deep learning model that automates and standardizes their localization in 3D fetal US volumes.

We evaluated our methods both qualitatively, through expert clinical review, and quantitatively. The proposed approach achieved a mean translation error of 3.21 $\pm$ 1.98mm and a mean rotation error of 5.31 $\pm$ 3.945$^\circ$ per plane, outperforming other state-of-the-art methods on 3D US volumes. Clinical assessments further confirmed the effectiveness of both GT++ and 3DFETUS, demonstrating statistically significant improvements in plane estimation accuracy.
\end{abstract}

\begin{graphicalabstract}
\includegraphics[width=0.99\textwidth]{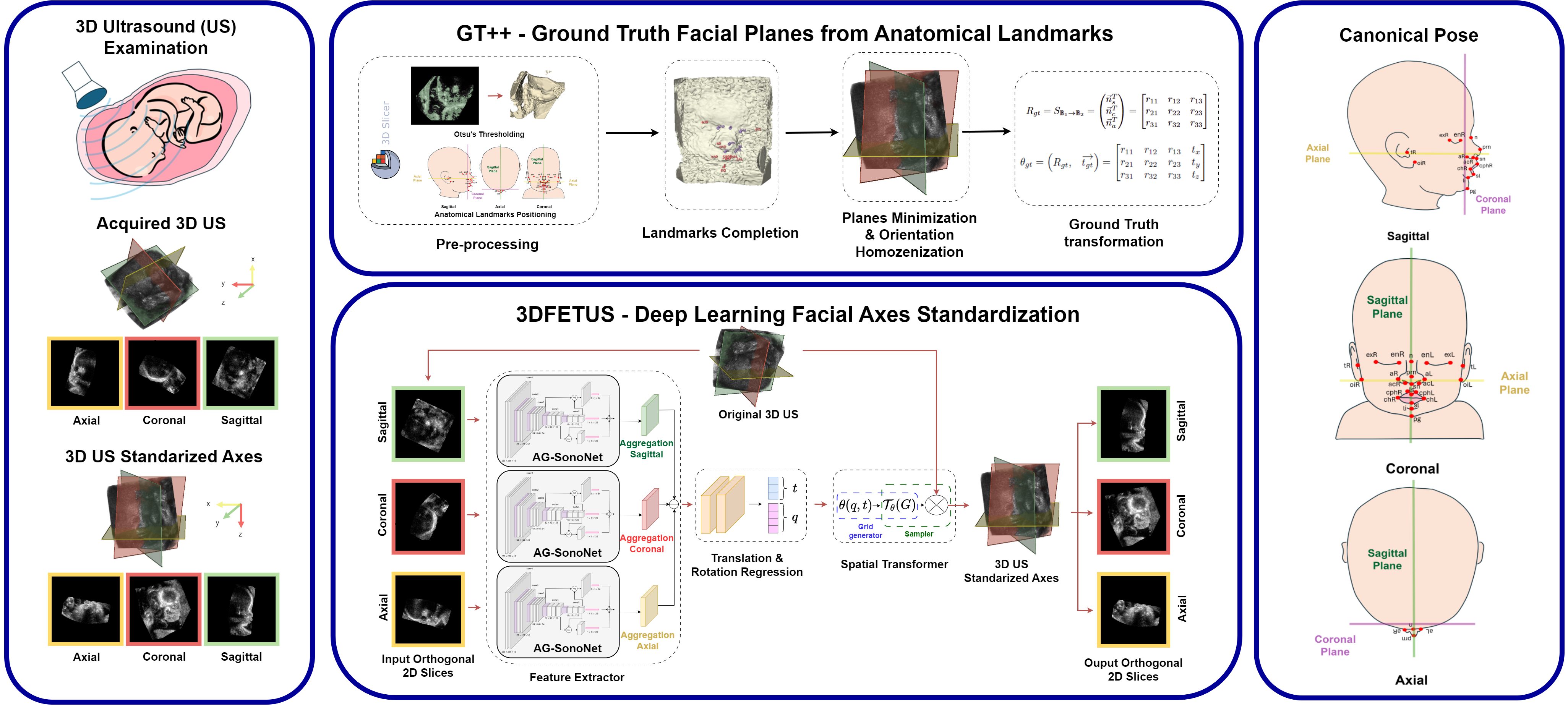}
\end{graphicalabstract}

\begin{highlights}
\item GT++: A novel benchmark for extracting ground-truth standard planes from 3D anatomical landmarks in fetal ultrasound volumes.
\item 3DFETUS: An automatic neural network for normalizing fetal facial orientation to a canonical frontal pose in 3D ultrasound volumes.
\item Clinical validation: Comprehensive evaluation of automated fetal facial analysis in a clinical setting.
\end{highlights}

\begin{keyword}
3D ultrasound \sep medical image standardization \sep anatomical plane localization \sep deep learning \sep pose normalization



\end{keyword}

\end{frontmatter}

\section{Introduction}
\label{sec:introduction}
The localization and standardization of anatomical planes in 3D medical imaging is a fundamental problem for quantitative analysis, visualization, and automated diagnosis. Variations in object pose, anatomy, and acquisition conditions introduce significant challenges, particularly in imaging modalities affected by noise and limited contrast. 

Ultrasound (US) is widely used due to its non-invasive, real-time, and cost-effective nature. From a computational perspective, US presents unique challenges, including speckle noise, acoustic shadowing, low contrast, and operator-dependent acquisition, making challenging automated plane localization. 3D US introduces additional pose variability, highlighting the need for robust plane standardization methods.

Fetal US provides an important real-world application of these challenges. It is routinely used to monitor fetal growth and anatomy throughout gestation. In clinical practice, 2D US remains the standard modality for routine assessment due to its accessibility and well-established protocols by international expert committees. 3D US offers richer volumetric information, but also introduces additional variability that underscores the need for robust, automated plane standardization methods.

Among the structures routinely evaluated, craniofacial anatomy is of particular interest, as face and skull abnormalities may indicate developmental disturbances or genetic syndromes \citep{Junaid2022,Bartzela2017,WADDINGTON2025}. To facilitate diagnosis, the International Society of Ultrasound in Obstetrics and Gynecology (ISUOG \footnote{https://www.isuog.org/}) has issued several guidelines for examining the face and skull during the first and second trimester using US \citep{Salomon2022,Salomon2011}. Accurate assessment requires imaging in multiple orientations to capture relevant features, such as the forehead, orbits, nose, and lips. The fetal face can be examined in sagittal (detecting forehead and nose dysmorphisms), axial (assessing orbits, interocular distance, and ocular diameter), and coronal planes (critical for evaluating mouth, lips, and nose). Each plane provides essential information for identifying facial abnormalities. 

To ensure consistency and reliability, sonographers acquire standard planes (SPs), predefined anatomical views that enable precise measurement and detection of anomalies. However, SP acquisition is time-consuming and highly dependent on the operator's expertise, leading to potential biases due to the extensive search space,  maternal body mass index, amount of amniotic fluid, fetal size, week of gestation, the variability in fetal orientation and movement, and differences in sonographer experience \citep{Nerea_2024,Sarris2012}. For example, ISUOG strongly recommends assessing the median facial profile, as it can help identify conditions such as cleft lip, micrognathia, and nasal bone anomalies. To diagnose these conditions, different facial makers are analyzed. The maxillary gap, facial maxillary angle, and palatine maxillary diameter aids in cleft detection and should be confirmed across multiple planes. Additionally, visualizing the orbits and retronasal triangle is particularly useful for screening anomalies such as Down syndrome, where nasal bones may be absent or hypoplastic  \citep{Mak2019,Benacerraf2019,Lamanna2023}.

3D US complements 2D imaging by providing volumetric reconstructions that allow retrospective examination of SPs, multiplanar visualization, and improved evaluation of complex craniofacial structures \citep{Dyson2000,Merz2017}. Despite these advantages, 3D US shares limitations with 2D US, including operator dependence, susceptibility to motion and equipment artifact, and the need for accurate SP extraction for meaningful analysis.

To streamline this process, machine learning and later deep learning (DL) methods have been proposed to automatically extract SPs from 3D US volumes or videos \citep{Liu2019, RamirezZegarra2023}. These approaches aim to reduce dependence on operator expertise, improve workflow efficiency, and standardize assessments across clinical settings \citep{Matthew2022,Espinoza2013,He2021}. However, they face several challenges, including subjective quality assessments, ambiguous plane definitions due to visually similar slices, and various artifacts such as shadows, tissue deformation, and probe variability \citep{Sarris2011, Skelton2021, Sarris2012}. Machine learning methods have attempted to address these issues using hand-crafted features or geometric constraints tailored to specific plane types, such as radial component models \citep{Kwitt2013,Maraci2014,Rahmatullah2012,Ni2014}. While useful in constrained scenarios, these techniques are limited by their reliance on low-level features, which fail to capture the full complexity and variability of fetal US volumes. In contrast, DL approaches offer greater flexibility by learning hierarchical representations directly from data. However, their effectiveness is strongly tied to the quality and consistency of the ground truth (GT) used during training. The scarcity of large, well-annotated datasets increases the risk of overfitting, an on-going challenge in medical DL applications. Therefore, establishing anatomically robust GT definitions and designing loss functions that reflect clinically meaningful alignment are critical for advancing DL-based SP detection.

This study proposes an automated pipeline that localizes fetal facial SPs in 3D US volumes and produces a spatially standardized 3D representation. The main contributions are:

\begin{itemize}
    \item \textbf{GT++ Benchmark for Standard Plane Detection:} We introduce GT++, a novel benchmark for extracting GT planes from 3D anatomical landmarks on fetal surface models derived from 3D US volumes. GT++ provides a consistent and objective framework for SP definition, significantly reducing inter-operator variability in clinical annotations.
    
    \item \textbf{3DFETUS: 3D Fetal Face Ultrasound Standardization:} We propose 3DFETUS, a differentiable neural network that estimates the  transformation required to normalize fetal facial orientation to a canonical frontal pose. This pose standardization reduces variability caused by fetal motion or acquisition angles, improving SP detection accuracy.
    
    \item \textbf{Grid Loss-Based Training Strategy:} We introduce a novel grid-based loss function integrated within a differentiable spatial transformation block. This strategy guides the network to learn anatomically plausible spatial arrangements across sagittal, coronal, and axial planes, promoting robust and interpretable SP predictions.
    \item \textbf{Clinical Assessment of Automated Fetal Facial Evaluation:} We evaluate the proposed pipeline through a qualitative clinical assessment, which provides preliminary evidence that GT++ and 3DFETUS produce consistent, interpretable, and clinically meaningful SPs. These results support the potential for future clinical validation and integration into routine practice as a lightweight, low-cost AI tool compatible with standard 3D US systems, contributing to more accessible and standardized fetal facial evaluation in clinical settings.
\end{itemize}

We propose a pipeline that uses annotated anatomical landmarks to define consistent GT planes and learn spatial transformations in 3D volumes. Demonstrated on fetal US, this approach provides a structured way to standardize volumetric data, which could potentially be adapted to other imaging contexts or organs where reliable plane localization is needed.

\section{Related Work}
\subsection{Standard Plane Localization}
Recent DL approaches for automatic detection of 2D SPs in US images have primarily framed the task as image classification using convolutional neural networks (CNNs) or recurrent neural networks \citep{Chen_2017,Baumgartner2017,Lei2015,Zhen2023}. While effective at identifying whether a 2D slice is a SP, these methods provide no guidance on how to correct non-SP slices.

Other approaches reformulate the task as a regression problem, estimating transformation parameters to localize SPs in 3D US volumes. \citet{Li2018_2} introduced an iterative transformation network that simultaneously classifies SP and regresses transformation parameters using a CNN. In a subsequent work, they proposed a similar architecture without the classification branch, incorporating two distinct loss functions that leverage geometric information and image data from the extracted planes \citep{Li2018}. \citet{Namburete2018} proposed a multi-task network jointly performing brain localization, segmentation, and alignment based on intracranial features. The network leverages shared feature learning across tasks to estimate the affine transformations for accurate brain alignment. \citet{Di_Vece_2024} trained a regression CNN on synthetic data and fine-tuned it on real data to predict the 6-DoF pose of arbitrary planes relative to the fetal brain.

Reinforcement learning has also been explored for SP localization \citep{Dou2019,Huang2020,Zou2022,Li2021}, formulating the task as a sequential decision-making problem. However, reinforcement learning methods often rely on accurate initialization and are limited by discrete action spaces. Recent works have explored alternatives, such as a diffusion-based model for SP localization \citep{Dou2025} and ScanAhead \citep{Men2025}, which uses transformers to predict fetal head planes from video sequences.

Interestingly, some methods developed for image registration also share the goal of estimating spatial transformations, that align a moving image to a reference image. Recent models such as TransMorph and TransMatch \citep{Chen2022,Chen2024,Balakrishnan2019} integrate differentiable transformation blocks into DL pipelines, enabling end-to-end optimization. However, these models typically operate on full 3D volumes and require a fixed reference image, making them computationally intensive and less flexible for real-time clinical use \citep{DEVOS2019128}.

To enable real-time integration into clinical US systems, we propose a lightweight, reference-free approach that eliminates the need for pre-registration, fixed reference images, or constrained transformation spaces. The model directly regresses the transformation that standardizes the input 3D US volume axes into a canonical fetal face coordinate system, defined by an up-right pose. Operating on three 2D orthogonal slices and guided by a differentiable grid loss, the network learns anatomical alignment efficiently, significantly reducing computational complexity compared to full 3D registration methods.

\subsection{Fetal Face Detection}

Several works have addressed fetal face detection and SP localization in 3D US, employing machine learning and DL techniques. For instance, \citet{Feng_2009} proposed a fast learning-based approach that combines constrained marginal space learning with 2D profile refinement to detect the fetal face and enhance its visualization through an automatic carving algorithm. \citet{Baumgartner2017} developed a CNN-based detection network capable of classifying 13 SPs in fetal 2D US, including key facial planes such as the coronal view of the lips and sagittal view of the face. \citet{Singh2021} introduced a DL pipeline for fetal face detection, segmentation, and landmark localization, demonstrating robust performance even in multiple-fetus pregnancies. Earlier methods such as \citet{Lei2015} employed RootSIFT features with Fisher Vector encoding and support vector machine classifiers for SP recognition, while \citet{Yu2016} leveraged deep CNNs for more fine-grained plane classification. More recently, \citet{Chen2020} applied a 3D region proposal network combined with landmark-centered bounding boxes and graph priors to improve landmark detection accuracy. Complementing these, \citet{Nerea_2024} 
evaluated manual facial landmarking across gestation, reporting low observer variability and insights into fetal facial growth.

Despite these advances, significant challenges remain. Notably, none of the existing works explicitly focus on the standardization of fetal facial axes, a crucial step for improving clinical evaluation and enabling consistent comparison of facial features. As \citet{Nerea_2024} demonstrated, accurately locating anatomical facial landmarks in 3D US is feasible, but could greatly benefit from removing pose variability. This highlights the importance of standardizing the fetal facial planes within 3D US as a foundation for more robust and clinically meaningful landmark detection.

\subsection{Training Strategy}
DL models for US SP detection are typically developed using either supervised or unsupervised learning approaches \citep{RamirezZegarra2023}. Supervised models, the most common type, rely on GT annotations provided by expert clinicians \citep{Baumgartner2017, Di_Vece_2024}. However, generating these labels can be time-consuming, requiring exhaustive search, pre-registration steps, or manual annotation of anatomical landmarks. Moreover, SP labels suffer from high inter- and intra-observer variability \citep{Sarris2011}. GT quality and registration accuracy have a significant impact on model performance \citep{DiVece2024, DiVece2022}. To address this, \citet{DiVece2022} showed that fiducial-based labeling aligned with global anatomy improves consistency and anatomical relevance.

Unsupervised methods, in contrast, bypass labeled data by exploiting intrinsic patterns or geometric constraints. These models often rely on handcrafted features such as histogram matching or SIFT descriptors to guide alignment \citep{Kwitt2013,Maraci2014,Rahmatullah2012,Ni2014}. While effective in controlled settings, they lack generalizability and require expert-driven feature design.

As an alternative, phantom and synthetic datasets, generated using physical models or generative models, are increasingly used to supplement real data when annotations are scarce \citep{Rathbun2021,Qiao2026,Duan2025}. However, domain gaps limit generalization to real clinical scans, and validating anatomical realism in synthetic images remains challenging.

To overcome these limitations, our method leverages anatomical landmarks to extract GT facial planes, taking advantage of their demonstrated high accuracy and low intra- and inter-observer variability \citep{Nerea_2024}. 
\begin{figure*}
\centering
\includegraphics[width=0.95\textwidth]{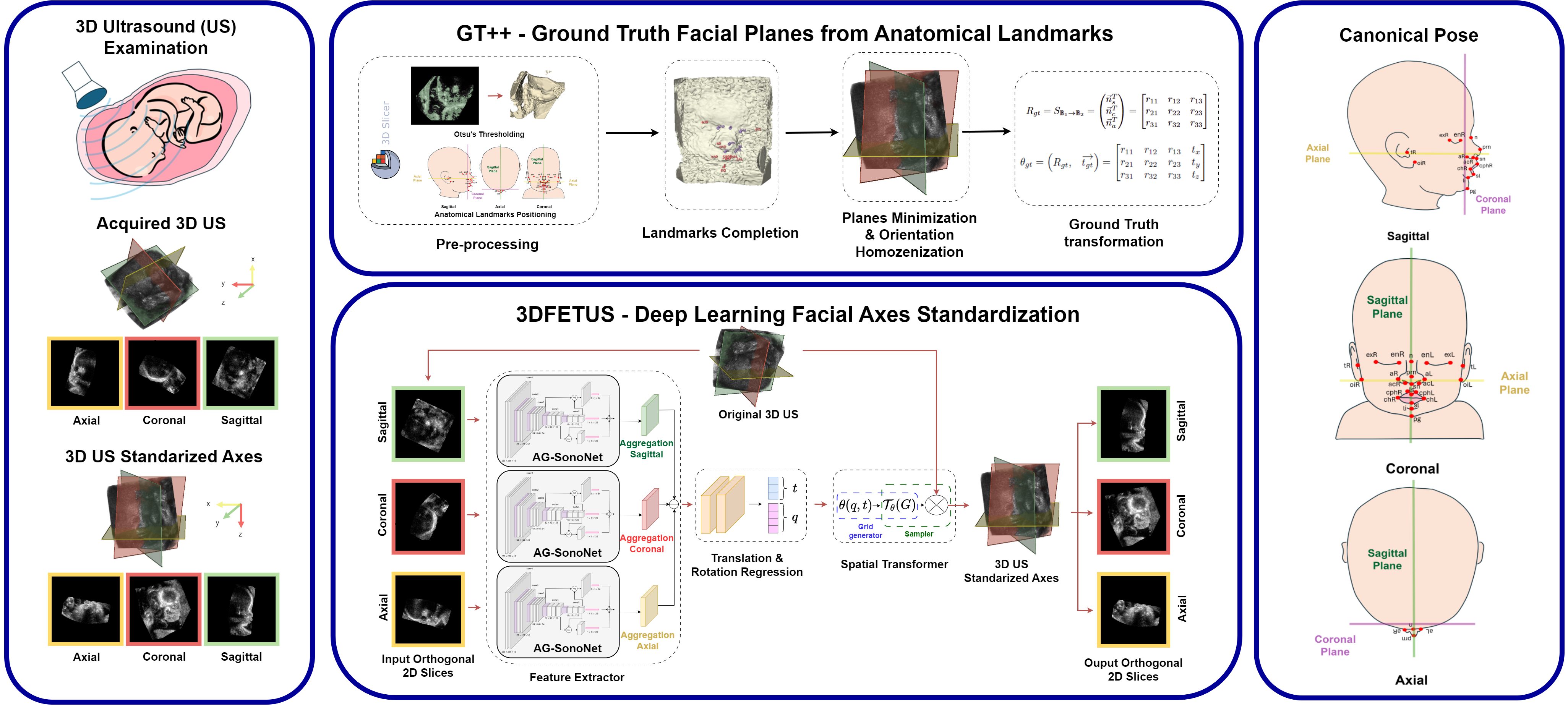}
\caption{Overview of the proposed 3DFETUS framework for automatic standardization of fetal facial planes in 3D US. The pipeline consists of two main stages: (1) GT++ Ground Truth Construction (top row), where the facial ground truth planes are derived from anatomical landmarks on 3D US volumes, and (2) 3DFETUS Network Architecture (bottom row), a deep learning model comprising three blocks: a feature extractor, affine transformation regressor (rotation and translation), and a spatial transformer module. The network processes three orthogonal US slices as input and predicts the transformation needed to align the fetal face into a canonical frontal pose. The spatial transformer then warps the original 3D volume accordingly, producing a standardized US volume and aligned sagittal, coronal, and axial facial planes. } \label{fig:overview}
\end{figure*}
\section{Data}
The dataset used consists of 973 fetal facial 3D US volumes from 133 subjects acquired between 20 and 35 weeks of gestation (mean 26.56 $\pm$ 2.72). All scans were obtained using a Voluson E8 RSA (BT-20) system equipped with with a convex probe (4D-RAB6-D, 2–8 MHz) at two hospitals in Barcelona (Hospital de Mar and Hospital Universitari Dexeus). Data collection was conducted in accordance with the approval of their Ethical Research Committee and the current legislation (Organic Law 15/1999). The study population comprises subjects from low-risk pregnancies, defined as those without any known pathology or family history of craniofacial or syndromic conditions, which were all carried to term. The dataset was divided into training, test and validation sets considering the gestational age (GA) and the subject independence (see Table \ref{tab:data_split} and Table \ref{tab:data_ga} for more detailed information). 

\begin{table}
\caption{Dataset split distribution for training, test, and validation.}\label{tab:data_split}
\centering
\begin{tabular}{|c|c|c|c|}
\cline{2-4}
\multicolumn{1}{c|}{}& \textbf{$\# $3D US}  & \textbf{$\%$} & \textbf{$\#$ subjects} \\
 \hline
\textbf{Training} & 629 & 65 & 83 \\
 \hline
\textbf{Test} & 147 & 15 & 27 \\
 \hline
\textbf{Validation} & 197 & 20 & 27 \\
 \hline
\end{tabular}
\end{table}

\begin{table*}
\caption{Gestational age and subject distribution of the dataset partition. w: weeks of gestation; $\#$: number; V: volumes; S: subjects.}
\centering
\begin{tabular}{|c|c|c|c|c|c|c|c|c|c|c|}
\cline{2-11}
\multicolumn{1}{c}{}& \multicolumn{2}{| c |}{\textbf{20-22w}} &  \multicolumn{2}{| c |}{\textbf{23-25w}} &  \multicolumn{2}{| c |}{\textbf{26-27w}} &  \multicolumn{2}{| c |}{\textbf{29-31w}} & \multicolumn{2}{| c |}{\textbf{32-35w}} \\
\cline{2-11}
 \multicolumn{1}{c|}{} & \textbf{$\#$ V} & \textbf{$\#$ S} & \textbf{$\#$ V} & \textbf{$\#$  S} & \textbf{$\#$ V}  & \textbf{$\#$ S}  & \textbf{$\# $ V} & \textbf{$\#$ S} & \textbf{$\#$ V}  & \textbf{$\#$ S} \\
 \hline
 \textbf{Train} & 71 & 30 & 44 & 7 & 397 & 68 & 87 & 12 & 24 & 10 \\
 \hline
 \textbf{Test} & 14 & 6 & 17 & 3 & 91 & 16 & 19 & 2 & 6 & 2\\
 \hline
 \textbf{Val} & 14 & 7 & 3 & 1 & 150 & 19 & 21 & 5 & 3 & 1 \\
 \hline
\end{tabular}\label{tab:data_ga}
\end{table*}

\section{ Ground Truth Facial Planes from Anatomical Landmarks}
The algorithm developed for extracting the GT transformation to standardize the US axes to the facial planes, based on anatomical landmarks, along with the resulting transformation and corresponding set of planes, will be referred to as GT++ throughout the paper. The pre-processig steps and the GT plane estimation was implemented using the open-source program 3D Slicer. All steps, except for the anatomical landmark positioning, were automated using Python scripts.

\subsection{Pre-processing}
\begin{figure}
\centering
\includegraphics[width=0.65\textwidth]{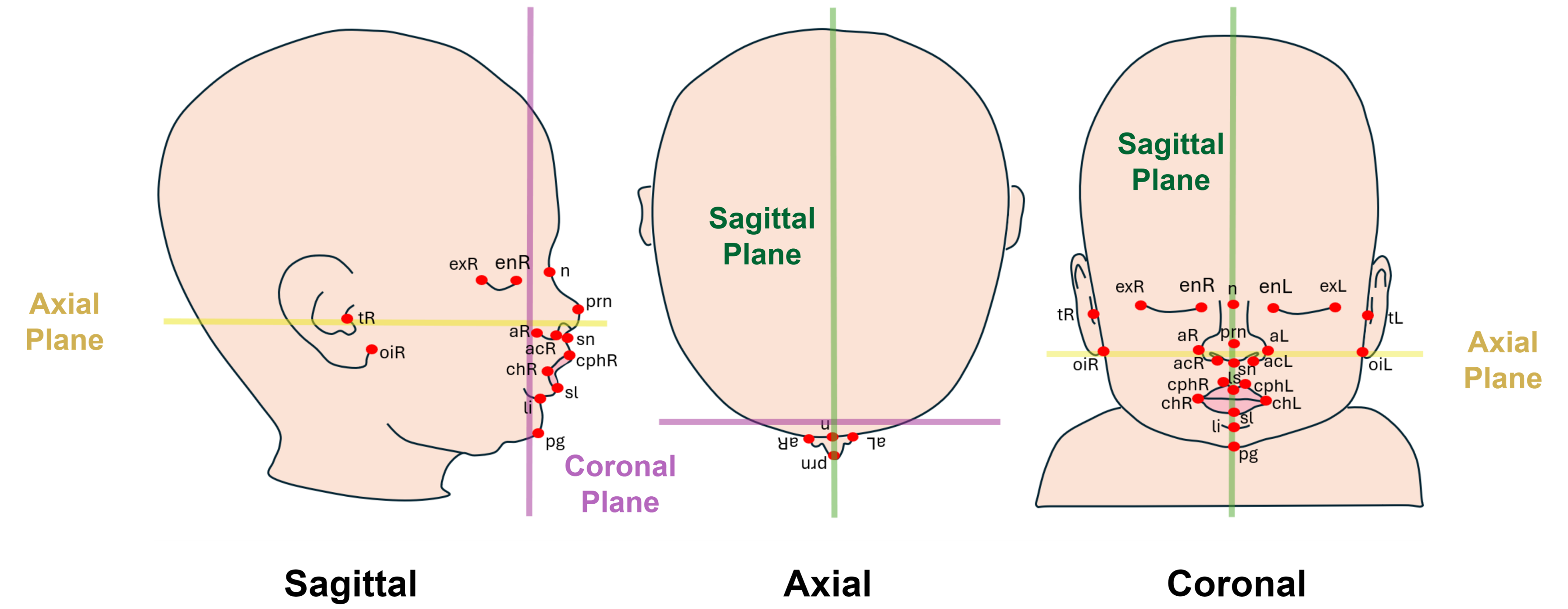}
\caption{Illustration of the anatomical landmarks, canonical facial pose and facial planes considered. Landmark abbreviations: Landmark abbreviations: exR, exL = exocanthion right, left; enR, enL = endocanthion right, left; n = nasion; aR, aL = alare right, left; acR, acL = alar crest right, left;
prn = pronasale; sn = subnasale; chR, chL = cheilion right, left; cphR, cphL = crista philtrum right, left; ls = labiale superius; li = labiale inferius; sl = sublabiale; pg = pogonion; tR, tL = tragion right, left; oiR, oiL = otobasion inferius right, left.} \label{fig:canonical_pose}
\end{figure}

The 3D US volumes are automatically segmented using the Otsu's thresholding method \citep{Otsu1979}. To reduce noise, only the largest connected component is retained, while smaller components are discarded. The resulting segmentation is then converted to a close surface representation and exported. Then, expert clinicians annotate the visible landmarks on the segmented surfaces from the 23 facial anatomical landmarks described in Figure \ref{fig:canonical_pose}. Landmark annotation is performed using the 3D Slicer Markups tool, where clinicians place a fiducial for each visible anatomical landmark. 

\subsection{Facial Planes Extraction} \label{sec:GT_extraction}

The GT facial axial, coronal and sagittal planes, which define the canonical axes of the face, are derived from the annotated landmarks on the 3D facial surfaces. In this way, we aim to achieve more objective GT planes by reducing subject-dependent variability, as the planes are estimated directly from the annotated landmarks rather than manually located. As demonstrated in \citep{Nerea_2024}, most facial landmarks can be located with high accuracy in 3D fetal facial US, exhibiting low overall intra- and inter-observer errors of 1.01 mm and 1.60 mm, respectively. By using the landmarks to define the facial planes, we enforce that our estimations are consistent and reliable across different individuals.

\begin{table}
\caption{Landmarks defining the canonical facial planes.}\label{tab:planes_lmks}
\centering
\begin{tabular}{|l|l|l|l|}
 \cline{2-4}
\multicolumn{1}{c|}{} &  \textbf{Sagittal} & \textbf{Coronal} & \textbf{Axial}\\
\hline
\multirow{2}{*}{Landmarks} & n, prn, sn, ls & chR, chL, enR, & aR, aL, acR, \\
 & li, sl, pg & enL, pg & acL, sn, prn\\
\hline
\end{tabular}
\end{table}

The initial step involves identifying the landmarks that define each plane. This process is guided by focus group guidelines, ISUOG standards, and the expertise of clinical professionals. The resulting decision criteria are detailed in Table \ref{tab:planes_lmks}. 
To define a plane, a minimum of three non-collinear points are required. If this condition is not met, the case is discarded, as the planes cannot be estimated.

\subsubsection{Landmark Completion}

To improve the accuracy and consistency of the estimated GT planes, we introduced a \textit{landmark completion} step. This aims to enhance the robustness of the plane estimation by inferring missing or unreliable landmarks, often affected by occlusion, limited field of view, or ultrasound-specific artifacts, using a statistical 3D morphable model (3DMM). Specifically, we constructed a
landmark-based 3DMM derived from the BabyFM framework \citep{Morales2025}, a model of infants facial anatomy. Since ears are typically not visible in fetal US, only the first 19 of the 23 facial landmarks defined in the model are used (see Figure~\ref{fig:canonical_pose}). The 3DMM is defined by a mean shape $\bar{x} \in \mathbb{R}^{3N}$ and a shape basis matrix $\Phi \in \mathbb{R}^{3N \times M}$, derived via Principal Component Analysis (PCA), where $N$ is the number of landmarks and $M$ is the number of retained modes of variation, defining the dimensionality of the model. Any landmark configuration can be approximately represented by:
\begin{equation} \label{eq:recon}
\vec{x} \approx \Phi \vec{\alpha} + \bar{x} \, ,
\end{equation}
where $\vec{x} \in \mathbb{R}^{3N}$ is the reconstructed shape and $\vec{\alpha}  \in \mathbb{R}^{M}$ are the shape parameters.

To estimate the missing landmarks, we first identify the visible landmarks and align them to the model mean using \textit{Procrustes analysis} \citep{Gower1975}. Following this alignment, we estimate the shape parameters $\hat{\alpha}$ by solving a constrained non-linear least-squares optimization problem. The objective consists of two terms. The first term measures the reconstruction error between the observed (visible) landmarks and their projection onto the morphable model subspace. The second term penalizes deviations from the distribution of plausible facial shapes, using the Mahalanobis distance of the shape parameters. Formally, we solve:
\begin{align} \label{eq:opt_corrected}
\hat{\alpha} = \arg \min_{\vec{\alpha}} \left\| \Phi_r \vec{\alpha} + \bar{x}_r - \vec{x_r} \right\|^2 + w_p \sum_{i=1}^M \frac{\alpha_i^2}{\lambda_i} 
\end{align}
where $\vec{x_r}$ and $\bar{x}_r$ are the observed and mean landmarks restricted to the visible subset, $\Phi_r$ is the corresponding submatrix of the shape basis, $\lambda_i$ are the eigenvalues of the PCA model and $w_p$ is the weight controlling the influence of the plausibility constraint. This formulation ensures that the reconstructed shape lies within a statistically plausible region of the morphable model space.

Once the optimal shape parameters $\hat{\alpha}$ are obtained, the full landmark configuration is reconstructed using Equation~\eqref{eq:recon}. Then, the \textit{inverse Procrustes transformation} maps the completed landmarks back to the original 3D US space, ensuring consistency with the original orientation and scale.

\subsubsection{Planes Minimization Function}
Each plane is defined by two key elements: a normal vector $\vec{n}  = (n_x,n_y,n_z)^T$, which indicates the direction perpendicular to the plane, and a centroid $\vec{c}  =(c_x,c_y,c_z)^T$, which represents the center point of the plane in 3D space. The normal vector sets the plane’s orientation, while the centroid anchors it at a specific location. Mathematically, any point $\vec{p}  =(x,y,z)^T$ that lies on the plane satisfies:
\begin{equation}  
\vec{n} \cdot (\vec{p} - \vec{c})= 0.  
\end{equation}
We want to estimate the three orthogonal facial planes that share the same center. This leads us to a least squares minimization problem with a total of 12 unknowns: the normal vectors of the sagittal, coronal, and axial planes $\vec{n}_s$, $\vec{n}_c$, $\vec{n}_a \in \mathbb{R}^3$, and the coordinates of their shared center $\vec{c} \in \mathbb{R}^3$. To ensure the desired geometric structure, the normal vectors are constrained to be orthonormal. Thus, the optimization problem can be formulated as:
\begin{align*}  
\text{min} \quad & f(\vec{\kappa}) =  w_{s} \cdot d_s(\vec{\kappa},{P_{s}}) + w_{c} \cdot d_c(\vec{\kappa},{P_{c}}) + w_{a} \cdot d_a(\vec{\kappa},{P_{a}}) \\
\text{subject to} \quad 
& \vec{n}_{i}^\top \vec{n}_{j} = 0 \quad \text{for } i \ne j,\; i,\; j \in \{a, c, s\} \\
& \|\vec{n}_{i}\|^2 = 1 \quad \text{for } i \in \{a, c, s\} 
\end{align*}  
Here, the optimization variable is $ \vec{\kappa} = [ \vec{n}_s^{T} \vec{n}_c^{T} \vec{n}_a^{T} \vec{c}^{T} ]^T \in \mathbb{R}^{12}$. The sets $P_{s}$, $ P_{c}$ and $ P_{a} \in \mathbb{R}^{3 \times L_j}$ are the visible landmarks for the sagittal, coronal, and axial planes, respectively (see Table~\ref{tab:planes_lmks}), and $w_s$, $w_c$, and $w_a$ are scalar weights controlling the contribution of each plane to the objective function. All the weights were set to $\frac{1}{3}$ so each plane contributes the same. The problem is solved using the Sequential Least Squares Programming (SLSQP) algorithm.

The distance function $d_j(\vec{\kappa}, P_j)$ measures the distance of landmarks to the candidate plane:
\begin{align*}  
d_j(\vec{\kappa}, P_j) = d(\vec{n_j}, \vec{c},P_j) = \sum^{L_j}_{i=1} |\vec{n}_j  \cdot(\vec{p}_{j,i} - \vec{c})|
\end{align*} 
where $\vec{p}_{j,i}$ is the $i$-th landmark on plane $j$, and $\vec{c}$ is the shared point through which all three planes pass.

The normal vectors are initialized using a random uniform distribution, while $ \vec{c}$ is initialized as the centroid point using the visible landmarks located on the scan.

\subsubsection{Homogenizing Normal Plane Orientation}
To remove ambiguities in the facial pose,
we enforce a consistent orientation for all plane normals relative to a front-right canonical pose (see Figure \ref{fig:canonical_pose}). Each annotated landmark $\vec{p^{'}}_i$ is projected onto the corresponding facial plane:
\begin{equation*}
   \vec{p^{'}}_i =  \vec{p_i} - \delta_i  \vec{n}  \quad
   \text{where} \quad
\delta_i =  ( \vec{p}_i -  \vec{c} ) \cdot\vec{n}
\end{equation*}  
with $\vec{c}$ denoting the plane's centroid and $\vec{n}$ the plane's normal. If the orientation of the projected landmark $\vec{p^{'}}$ does not match the predefined canonical orientation (as specified in 
Table \ref{tab:direction_normals}), the normal is flipped: $\vec{n} \leftarrow - \vec{n}.$

This procedure ensures that all normal vectors follow the established canonical orientation, maintaining uniformity across the facial surfaces.

\begin{table}
\caption{Homogenization of the planes normal orientation.}\label{tab:direction_normals}
\centering
\begin{tabular}{|l|l|l|l|}
 \cline{2-4}
\multicolumn{1}{c|}{} &  \textbf{Sagittal} & \textbf{Coronal} & \textbf{Axial}\\
\hline
\multirow{2}{*}{Positive}& enR, exR, aR, ls, & li, cphR, cphL, & enR, exR, n,\\
& acR, chR &prn, sn, acL, acR & enL, exL\\
\hline
\multirow{2}{*}{Negative} & enL, exL, aL, &  exR, exL & pg, sl, li, chR, chL,\\
 & acL, chL &  & ls, cphL, cphR\\
\hline
\end{tabular}
\end{table}

\subsubsection{Ground truth transformation}
The extracted GT normal vectors of the three orthogonal anatomical planes are used to compute the rotation matrix $R_{gt} \in \mathbb{R}^{3\times3}$  that standardizes the volume axes to the estimated facial SPs. This matrix can be interpreted as the change-of-basis matrix $S_{\mathbb{B}_1 \rightarrow \mathbb{B}_2 }$, where $\mathbb{B}_1$ denotes the original axes of the acquired volume (canonical basis $\in \mathbb{R}^{3}$) and $\mathbb{B}_2$ the orthonormal basis defined by the GT normal vectors (facial anatomical basis). Under the assumption that the normal vectors $ \vec{n}_s, \vec{n}_c, \vec{n}_a $ form a right-handed orthonormal basis, $S_{\mathbb{B}_1 \rightarrow \mathbb{B}_2 }$ is equivalent to a rotation matrix, and can be written as:

\begin{equation}\label{eq:rotation_gt}
    R_{gt}= S_{\mathbb{B}_1 \rightarrow \mathbb{B}_2 } =  \begin{pmatrix}
        \vec{n}_{s}^T \\\vec{n}_{c}^T \\ \vec{n}_{a}^T 
    \end{pmatrix} = \begin{bmatrix}
    r_{11} & r_{12} & r_{13} \\
    r_{21} & r_{22} & r_{23} \\
    r_{31} & r_{32} & r_{33} \\
\end{bmatrix}
\end{equation}



For rotation regression, $R_{gt}$ is converted to a unit quaternion representation $\vec{q_{gt}} = (q_0, q_1, q_2, q_3)^{T}$ (see Section \ref{sec:regression}) given by:
\begin{equation}\label{eq:quaternion_gt}
    \vec{q_{gt}} = \begin{pmatrix}
    q_0\\ q_1\\ q_2 \\ q_3\\ \end{pmatrix}  \\ =\begin{bmatrix}
    \frac{1}{2} \sqrt{1 + r_{11} - r_{22} - r_{33}}\\
    \frac{r_{12} + r_{21}}{4 q_0}\\
    \frac{r_{13} + r_{31}}{4 q_0}\\
    \frac{r_{23} - r_{32}}{4 q_0}\\
\end{bmatrix}  
\end{equation}
The center of the planes is the GT translation $\vec{t_{gt}} = \vec{c} \in \mathbb{R}^{3}$. Then, to perform the training the GT transformation matrix is expressed as:
\begin{equation}
    \theta_{gt} \begin{pmatrix}
    \vec{q_{gt}}, & \vec{t_{gt}} \\ 
\end{pmatrix} =\begin{bmatrix}
    r_{11} & r_{12} & r_{13} & t_x\\
    r_{21} & r_{22} & r_{23} & t_y\\
    r_{31} & r_{32} & r_{33} & t_z\\
\end{bmatrix}  \in \mathbb{R}^{3\times4}
\label{eq_gt_theta}
\end{equation} 
To ensure compatibility with the spatial transformer block (see Section \ref{sec:spatial}), the translation is expressed in image relative size, $\vec{t_{gt}}$ is in the range $[-1,1]$. The 2D GT sagittal, coronal and axial SPs are obtained as $I_{gt} = [I_s, I_c, I_a]$ where $I_s = V_{gt}(1,\frac{H}{2}, : , :)$, $I_c =  V_{gt}(1,:, \frac{W}{2} , :) $ and $I_a =  V_{gt}(1,:, : , \frac{D}{2})$, and $ V_{gt}$ is the transformed US volume using $\theta_{gt}$.

\label{sec:methods}

\section{Deep learning facial axes standardization}
To automate the standardization process, we introduce the 3D Fetal Face Ultrasound Standardization (3DFETUS) architecture. Using the obtained GT++ transformation (which align the 3D US axes to the facial SPs) as training supervision, the network learns to automatically standardize the fetal facial orientation to the canonical front-right pose. Given a randomly oriented input volume, 3DFETUS outputs its standardized version (see Figure \ref{fig:overview}). It consists of three main blocks: 1) Feature Extractor, which takes as input the orthogonal planes and extracts the relevant features; 2) Rotation and Translation Regression, which utilizes the extracted features to estimate rotation and translation transformations to align the facial axes to the standard position; 3) Spatial Transformer Block, which is a differentiable block that applies the estimated transformation to the original 3D US.

\subsection{Pre-processing}
The input 3D US volumes are pre-processed to a standardized size of $256\times256\times 256$ to facilitate network training. First, the images are down-sampled by a factor of two to reduce computational cost. Next, symmetric zero-padding is applied around the image center to obtain a volume $U \in \mathbb{R}^{C\times H \times W \times D}$ where $H,W,D = 256$ are the height, width, and depth dimension and $C= 1$. The latter is performed to ensure that no information is cropped-out during subsequent rotation and translation. The 2D initial planes are defined by $I_{0} = [I_1, I_2, I_3]$ with $I_1 = U(1,\frac{H}{2}, : , :)$, $I_2 = U(1,:, \frac{W}{2} , :) $ and $I_3 = U(1,:, : , \frac{D}{2})$, corresponding to the acquisition planes at the center of the 3D US volume.

To improve the performance and generalization of the network to unseen data, data augmentation is applied to the training volumes, including random geometric transformations (e.g., rotations and translations) and intensity variations. 

\subsection{Feature Extractor Block}

The inputs to the feature extractor block ($I_0 \in \mathbb{R}^{H\times W \times 3} $) consist of the three orthogonal planes located at the center of the 3D US volume. Each branch employs the AG-SonoNet \citep{Schlemper_2019} as the backbone feature extractor, with shared weights across branches. Each view incorporates a specialized aggregation block that integrates attention information from multiple layers of the network to extract view-specific features. The aggregated features from all branches are then concatenated and passed to the translation and rotation regression block.

\subsection{Translation \& Rotation Regression Block} \label{sec:regression}

It consists of two fully connected layers that convert the extracted features from the three orthogonal planes into the translation and rotation necessary to achieve the standardized axes/planes. The output is a regression vector $ \vec{\tau} =[\vec{q_{es}}, \vec{t_{es}}] \in \mathbb{R}^7$. The first four elements correspond to the quaternion representation of the rotation matrix, $\vec{q_{es}}= (q_0,q_1,q_2,q_3)^{T}$. To ensure a valid rotation, the quaternion is normalized such that $||\vec{q_{es}}||=1$, which is enforced by a normalization layer after the last fully connected layer. Given $\vec{q_{es}}$, the estimated rotation can be expressed as rotation matrix $R_{es} \in \mathbb{R}^{3 \times 3}$, analogous to the conversion shown in Equation \ref{eq:rotation_gt} and \ref{eq:quaternion_gt} for the GT rotation. The remaining three elements correspond to the translation vector $\vec{t_{es}}= (t_x,t_y,t_z)^{T}$ where each component is within the range $[-1,1]$.
\subsection{Spatial Transformer Block} \label{sec:spatial}
This differentiable module applies the learned 3D transformation to the input US volume $U \in R^{C \times H \times W \times D}$ to generate the standardized volume $V_{es} \in R^{C \times H \times W \times D}$. This is achieved following the Spatial Transformer Networks framework \citep{Jaderberg_2015}, leveraging differentiable grid generation and trilinear interpolation. 

The module defines a regular output grid with normalized coordinates in $[-1,1]$ range:
\begin{equation}
G = \{\vec{g_i}=(x^{\mathrm{out}}_i,\,y^{\mathrm{out}}_i,\,z^{\mathrm{out}}_i)^{T}\}.
\end{equation}
Each point $\vec{g_i}$ is mapped into the input volume using the learned affine transformation:
\begin{align}
\begin{pmatrix}
    x^{inp}_i \\
    y^{inp}_i \\
    z^{inp}_i \\
\end{pmatrix} &=
\mathcal{T}_{\theta_{es}}(\vec{g_i}) = \theta_{es}(\vec{q_{es}},\vec{t_{es}}) \begin{pmatrix}
    x^{out}_i \\
    y^{out}_i \\
    z^{out}_i \\
    1\\
\end{pmatrix}\label{eq_spatial_transform}
\end{align}
where $\theta_{es}(q_{es},t_{es})$ represents the learned transformation parameters and $x^{\mathrm{inp}}_i,\,y^{\mathrm{inp}}_i,\,z^{\mathrm{inp}}_i$ are the coordinates of the input volume $U$. Using the sampling grid, each output voxel $\vec{v_i}$ is computed from the input volume $U$ through trilinear interpolation:
\begin{equation}
\vec{v_i} = \sum_{n=1}^{H} \sum_{m=1}^{W} \sum_{l=1}^{D} U_{n,m,l}\,
w\bigl(x_i^{\mathrm{inp}} - m\bigr)\,
w\bigl(y_i^{\mathrm{inp}} - n\bigr)\,
w\bigl(z_i^{\mathrm{inp}} - l\bigr),
\end{equation}
with the interpolation weights defined as:
\begin{equation}
w(u) = \max\bigl(0,\;1 - |u|\bigr).
\end{equation}
These weights ensure that only the immediate neighbors of the sampling point contribute, while preserving smooth differentiability for backpropagation.

After transforming $U$ to $V_{es}$, the resulting facial SPs are extracted as $I_{es} = [I_s, I_c, I_a]$ where $I_s = V_{es}(1,\frac{H}{2}, : , :)$, $I_c = V_{es}(1,:, \frac{W}{2} , :) $ and $I_a = V_{es}(1,:, : , \frac{D}{2})$.

\subsection{Cumulative Transformations \& Initialization}
At initialization ($it=0$), the volume is transformed using a random rotation $R_0$ defined by the Euclidean angles $\zeta_x, \zeta_y, \zeta_z \in [-20,20]$ degrees and a random translation $\vec{t_0}  =(t_x,t_y,t_z)^{T}$ in the range $[-0.05, 0.05]$. To improve accuracy and preserve image quality, the network can refine the predicted transformation over multiple iterations. At each iteration $it$, the network estimates a residual rotation and translation ($R_{es}$, $t_{es}$) that incrementally aligns the volume to the standardized frontal-right pose. The accumulated estimated transformation is updated as $R_{es}^{it} = R_{es}^{it-1} R_{es}$ and $\vec{t_{es}^{it}}= \vec{t_{es}^{it-1}} + \vec{t_{es}}$, while the corresponding GT transformation is updated relative to the accumulated estimate $R_{gt}^{it} = R_{es}^{-1} R_{gt}^{it-1}$ and $\vec{t_{gt}^{it}} = - (R_{es}^{it-1})^{-1} \vec{t_{es}} + (R_{es}^{it-1})^{-1} \vec{t_{gt}}$. This iterative approach allows the network to progressively refine the alignment while maintaining a single cumulative transformation applied to the input volume, avoiding repeated resampling and preserving image quality. We found that 3 iterations are enough to improve performance by refining the SPs estimates. Notably, although the transformations are refined over multiple iterations at inference time, the network itself is trained in a single pass, without requiring iterative training.

\subsection{Network Loss} 
A key challenge in 6D pose estimation lies in selecting appropriate loss functions and representation spaces for regressing affine transformations, as the optimal approach remains unresolved \citep{Zhou2019, Alvarez-Tun2023}. While translation is typically modeled with three parameters, rotation, often represented by Euler angles or quaternions, poses greater difficulty due to its non-Euclidean and discontinuous nature. To overcome this, many works recommend operating within the SO(3) manifold. More recently, grid-based pose representations have been proposed as a stable and accurate alternative for training deep networks \citep{Park2022}. Therefore, to train the 3D facial axes standardization, we use the Mean Absolute Error (MAE) between the sampling grid using predicted transformation parameters \(\theta_{es}\) versus the one obtained using the GT transformation parameters \(\theta_{gt}\):  
\begin{equation}
\mathcal{L} (\theta_{es}, \theta_{gt}) = \frac{1}{N} \sum_{i=1}^{N} |\mathcal{T}_{\theta_{es}}(\vec{g_i}) - \mathcal{T}_{\theta_{gt}^{-1}}(\vec{g_i})|   
\end{equation}
where $\mathcal{T}_{\theta_{es}}(\vec{g_i})$ represents the transformed grid generated by the predicted parameters $\theta_{es}$ (see Equation \ref{eq_spatial_transform}), whereas $\mathcal{T}_{\theta_{gt}^{-1}}(\vec{g_i})$ represents the transformed grid generated by the inverse of the GT parameters $\theta_{gt}$ (see Equation \ref{eq_gt_theta}). This approach promotes a more straightforward implementation while still ensuring that both rotation and translation are effectively captured in the learning process.

By using a single MAE loss function to capture the discrepancies caused by both rotation and translation estimated with respect to the GT, we avoid the need for separate hyper-parameters for each loss term (rotation and translation). This simplification is particularly useful in this scenario where tuning can become complex and computationally expensive.




\section{Experiments}
To evaluate the proposed methodology, we conduct both quantitative and qualitative analyses of the predicted SPs, assessing both the GT estimation and the performance of the DL model. Specifically, we evaluate: (1) the GT annotations, comparing a baseline version without landmark completion (GT-base) to an enhanced version with the completion step (GT++), and (2) the performance of our model, 3DFETUS, in comparison to a state-of-the art method.

\subsection{Metrics}
\textbf{Qualitative score}: Denoted by $z$, this score captures the subjective assessment of expert clinicians about SP quality, on a 1–5 scale (1 = lowest, 5 = highest), based on rotational and translational deviations from an ideal SP.

\textbf{Quantitative Metrics}:
The quantitative evaluation assesses discrepancies between two plane sets, such as different GT versions (e.g., with vs. without landmarks completion) or between model predictions and GT++. The metrics include:
\begin{itemize}
 \item \textit{Rotation error}: using two complementary metrics:
  \begin{itemize}
    \item \textit{Geodesic distance on} $\mathrm{SO}(3)$: measures the overall angular deviation between plane orientations in 3D space, capturing all rotational components jointly.
    \item \textit{Plane-wise angular error}: Calculates the angle between the normal vectors of the two planes for each view, providing a per-plane discrepancy.
  \end{itemize}
      
    \item \textit{Translation error}: computed as the mean Euclidean distance between the centers of the two compared planes.
\end{itemize}
\textbf{Statistical analysis}:
A paired t-test was used to determine whether differences between the two model groups were statistically significant. Additionally, a one-way ANOVA test was performed to evaluate whether model performance varied significantly across different GA ranges. For both tests, statistical significance was set at p $\leq$ 0.01, indicating that observed differences are unlikely to arise by chance.

\subsection{Evaluation of the proposed GT plane extraction}
\label{sec:quality_clinical}
\textbf{Qualitative clinical assessment}:
To assess the quality of the proposed GT extraction, we conducted a comprehensive clinical evaluation in which experienced clinicians independently reviewed the facial SP obtained with and without landmarks completion step —referred to as GT++ and GT-base, respectively. Six expert obstetricians qualitatively assessed the 2D SPs derived from the test set of 147 3D US volumes from 27 different subjects. Each clinician rated the quality score ($z$) of the images generated with GT++ and GT-base. For each 3D US volume, clinicians independently evaluated the axial, sagittal, and coronal planes. To mitigate potential bias in the evaluation process, the presentation order of the images was randomized. The dataset was divided into six distinct, non-overlapping splits, with each clinician assigned three splits at random, ensuring that each split was ultimately evaluated by three different clinicians. 

\textbf{Inter-observer variability}:
Because the GT is derived from manually annotated landmarks, variability in these annotations can impact the obtained GT. Therefore, to assess the robustness of our GT++ facial plane extraction algorithm, we evaluated the error resulting from inter-observer variations in the positioning of facial landmarks. To this end, we used landmark annotations from 3 clinical observers on 30 3D facial surfaces with and without the step of the landmarks completion (i.e., GT++ vs. GT-base). The analysis is performed on the same subset analyzed in \citep{Nerea_2024}.

\subsection{Evaluation of the deep learning model}
\textbf{Qualitative clinical assessment}:
To assess the quality of the resulting model, we compared 3DFETUS against the methodology described by \citet{Li2018}. We conducted a comprehensive clinical quality evaluation following the same procedure described in Section \ref{sec:quality_clinical}.

\textbf{Performance analysis}:
The 3DFETUS performance was compared against three baselines: the upper bound, defined by the acquisition qualitative errors between the original US acquisition and the estimated GT++; the lower bound, given by the inter-observer variability of the GT++ annotations; and the state of the art method Iterative Transformation Network (ITN) \citep{Li2018}, trained on the same data. Comparisons are based on both rotation and translation errors between estimated planes and corresponding GT references.

\textbf{Gestational Age analysis}:
To assess the impact of GA on plane standardization, we evaluated the performance of 3DFETUS across different GA ranges in the test set, as defined in the dataset description. To determine whether performance differences among GA groups were statistically significant, we conducted a one-way ANOVA test. Statistical significance was defined as a p-value $< 0.01$.

\textbf{Ablation Loss}:
To analyze the impact of using a grid loss, which incorporates rotation and translation simultaneously, we compared it against using an aggregated loss that combines rotation, translation and image losses in separate terms.

\subsection{Quantitative-Qualitative relation}
We analyzed the relationship between translation and rotation errors and clinical quality scores by studying the data distribution within a continuous space. To achieve a denser and smoother representation, Gaussian splatting interpolation was applied, allowing a clearer visualization of how qualitative clinical assessments relate to quantitative results.

\subsection{Implementation details}
All models were trained and evaluated using a single GPU (NVIDIA Quadro:2(S:0), 6 GB VRAM). Training was conducted with a batch size of 24, a learning rate of 0.0001, a downsampling factor of 2, and an input image size of 256×256. The ITN \cite{Li2018} model, with approximately 70.8 million parameters, achieved an average inference time of 5.8 seconds, while the lighter 3DFETUS model, with 1.15 million parameters, had a slightly lower inference time of 5.28 seconds. 

\section{Results}

\begin{figure}
\centering
\includegraphics[width=0.8 \textwidth]{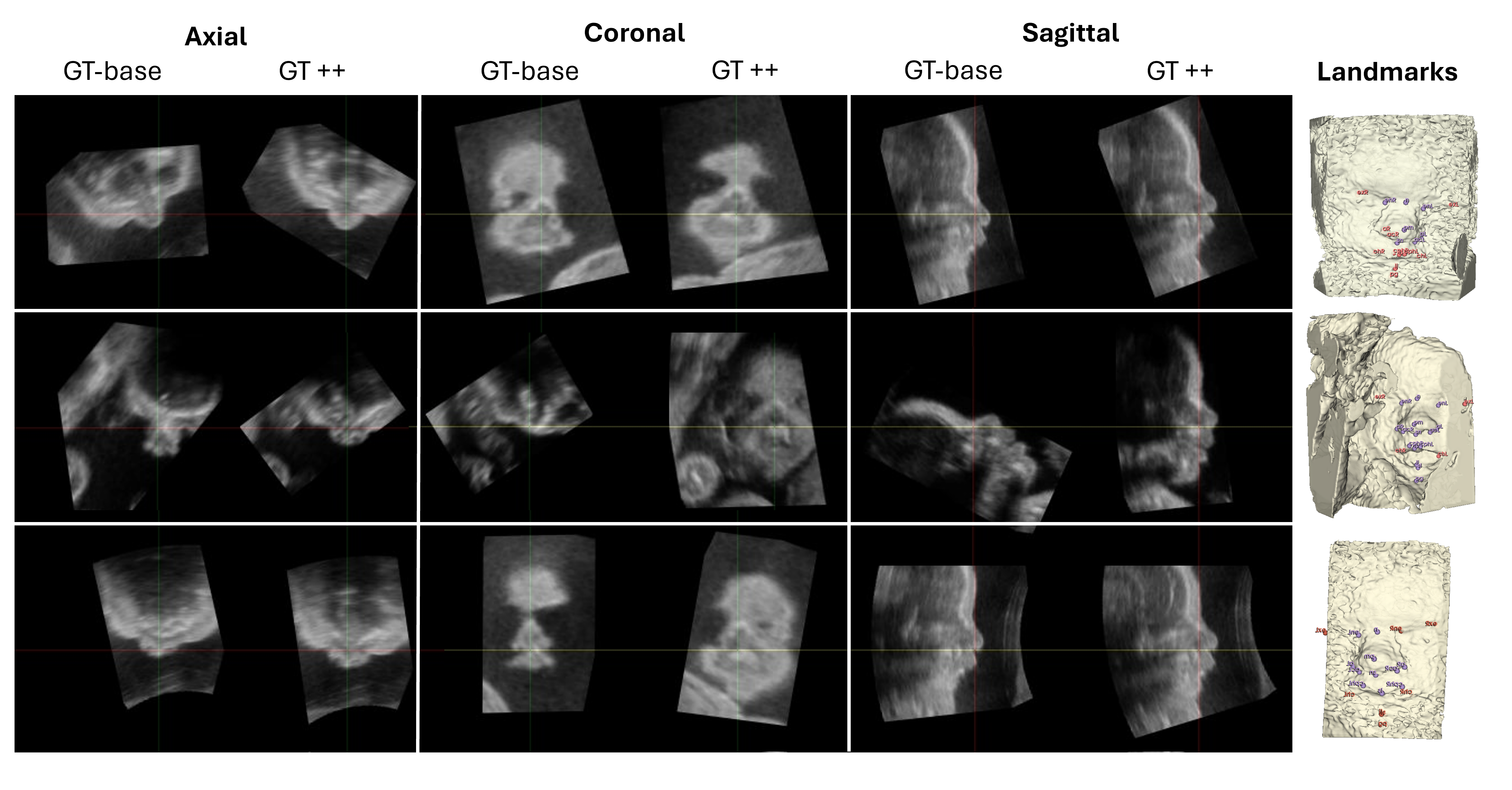} 
\vspace{-3mm}
\caption{Examples of facial planes generated using GT-base and GT++. Clinician-defined landmarks are shown in purple, while red landmarks represent those added through the completion process. Each row corresponds to a different subject. The columns display sagittal, coronal, and axial views, for both methodologies to facilitate comparison.} \label{fig:gt_comparison}
\end{figure} 
\subsection{Evaluation of the proposed GT plane extraction}
\begin{table}
\centering
\caption{Qualitative scores of the GT facial plane (mean$\pm$SD). Indicated with an * (paired t-test $p < 0.01$). GT-base: without landmarks completion; GT++: with landmarks completion. }  \label{tb: score_gt}
\begin{tabular}{|l|l|l|l|l|}
\hline
\textbf{Model} &\textbf{Global} & \textbf{Sagittal} & \textbf{Coronal} &\textbf{Axial}\\
\hline
GT-base & 3.49 $\pm$ 1.06 & 3.88$\pm$ 1.07 & 3.13$\pm$ 1.01 & 3.47 $\pm$ 0.96\\
\hline
GT++ & 4.06$\pm$ 0.57$^{*}$ & 4.41$\pm$ 0.46$^{*}$ & 3.98$\pm$ 0.53$^{*}$ & 3.79$\pm$ 0.57$^{*}$ \\
\hline
\end{tabular}
\end{table}
\textbf{Clinicians Quality Score}:
As shown in Table \ref{tb: score_gt}, the GT++ planes achieved a higher mean quality score ($p < 0.01$), indicating improved clinical acceptability of the facial SP. Considering a quality score threshold of 3 or higher as acceptable/good, the proposed GT++ method achieved good facial plane standardization results in 96\% of the test set, with only 4\% classified as low or bad quality. In comparison, the GT-base without the landmarks completion obtained 77\% good-quality results and 23\% low quality. These findings indicate that the proposed GT++ including the completion of the missing landmarks improves the proportion of acceptable quality scans by 19\% with respect to only using the annotated landmarks by clinicians.

\textbf{Inter-observer variability}:
The results, presented in Table \ref{tb: inter_obs_error}, demonstrate that GT++ significantly reduces both translation and rotation inter-observer error. These findings suggest that the proposed GT++ methodology enhances the consistency and robustness of the GT annotation. By completing missing landmark information using statistical priors from newborn faces (as described in Section \ref{sec:GT_extraction}), the estimated planes become more robust to inter-observer variability and missing landmarks caused by occlusions or noise in fetal facial region.
\begin{table}
\centering
\caption{GT errors per plane due to landmark variability (mean$\pm$SD). Indicated with an * (paired t-test $p < 0.01$). GT-base: without landmarks completion; GT++: with landmarks completion. } \label{tb: inter_obs_error} 
\begin{tabular}{|l|l|l|l|l|}
\hline
\textbf{Metric} &\textbf{Model} & \textbf{Sagittal} & \textbf{Axial} &\textbf{Coronal}\\
\hline
$t$ & GT-base & 1.60 $\pm$ 1.74 & 1.94 $\pm$ 1.84 & 1.51 $\pm$ 1.40\\
\cline{2-5}
(mm) & GT++ & 0.98 $\pm$ 1.13$^{*}$ & 1.14$\pm$ 1.26$^{*}$ & 1.08 $\pm$ 1.30$^{*}$\\
\hline
$R$ & GT-base & 11.35 $\pm$ 21.66 &  7.08 $\pm$ 9.60 &  12.63 $\pm$ 26.73\\
\cline{2-5}
($^{\circ}$)& GT++ & 5.06 $\pm$ 5.41$^{*}$ & 4.67 $\pm$ 5.23$^{*}$ & 4.11 $\pm$ 4.49$^{*}$\\
\hline
\end{tabular}
\end{table}

Figure \ref{fig:gt_comparison} provides qualitative examples comparing GT-base and GT++ facial planes, illustrating how landmark completion leads to more accurate corrections and improvements in the predicted planes. 

\subsection{Evaluation of the deep learning model}

\begin{table}
\caption{Quantitative evaluation of the 3D standardization of fetal volumes on the test set. Geodesic rotation error (SO(3)), mean Euclidean angle error (EA), and mean absolute translation error (t) are computed across all three planes, providing a global measure of how well the 3D volume has been aligned to the canonical upright pose.}\label{tab:global_error}
\vspace{4mm}
\centering
\begin{tabular}{|l|l|l|l|}
\hline
\textbf{Method} &  \textbf{SO(3) ($^{\circ}$) } $\downarrow$  & \textbf{EA($^{\circ}$) } $\downarrow$  & \textbf{t (mm)} $\downarrow$ \\ 
\hline
Acquisition &  116.14 $\pm$ 28.89 & 54.29 $\pm$ 26.69 & 
20.45 $\pm$ 6.10\\ 
\hline
Inter-observer &  21.91$\pm$34.38 & 10.35$\pm$19.33 & 3.31$\pm$2.46\\ 
\hline
ITN \citep{Li2018} & 34.43$\pm$46.40 & 21.13$\pm$37.25 & 14.51$\pm$ 23.81 \\
\hline
3DFETUS  & \textbf{10.72$\pm$6.39} & \textbf{10.61$\pm$7.48} & \textbf{6.78$\pm$3.89} \\ 
\hline
\end{tabular}
\end{table}

\begin{figure}
\centering
\includegraphics[width=0.85\textwidth]{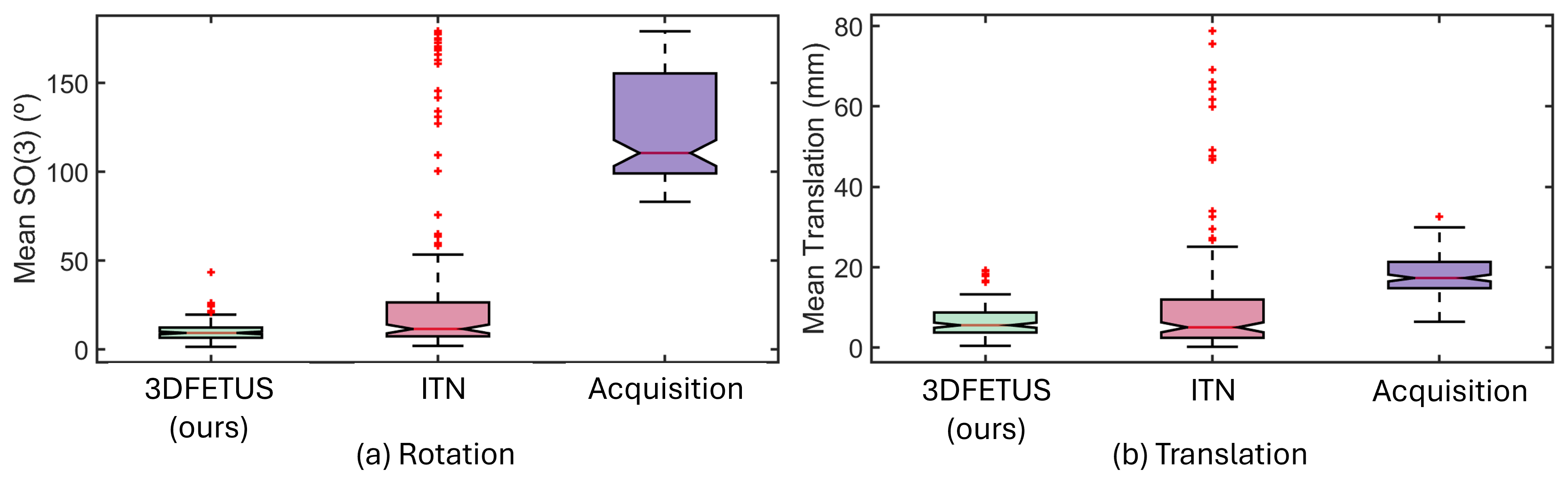}
\caption{Comparison of test set examples using 3DFETUS and ITN \citep{Li2018}. For each example, the Euclidean angle (EA, in degrees) and translation error ($t$, in mm) per plane are reported with respect to the ground truth (GT). The mean quality score ($z$), as rated by clinicians, is also provided.} \label{fig:sota_3dfetus_comp}
\end{figure}

\textbf{Performance analysis}:
\begin{figure}
\centering
\includegraphics[width=0.85\textwidth]{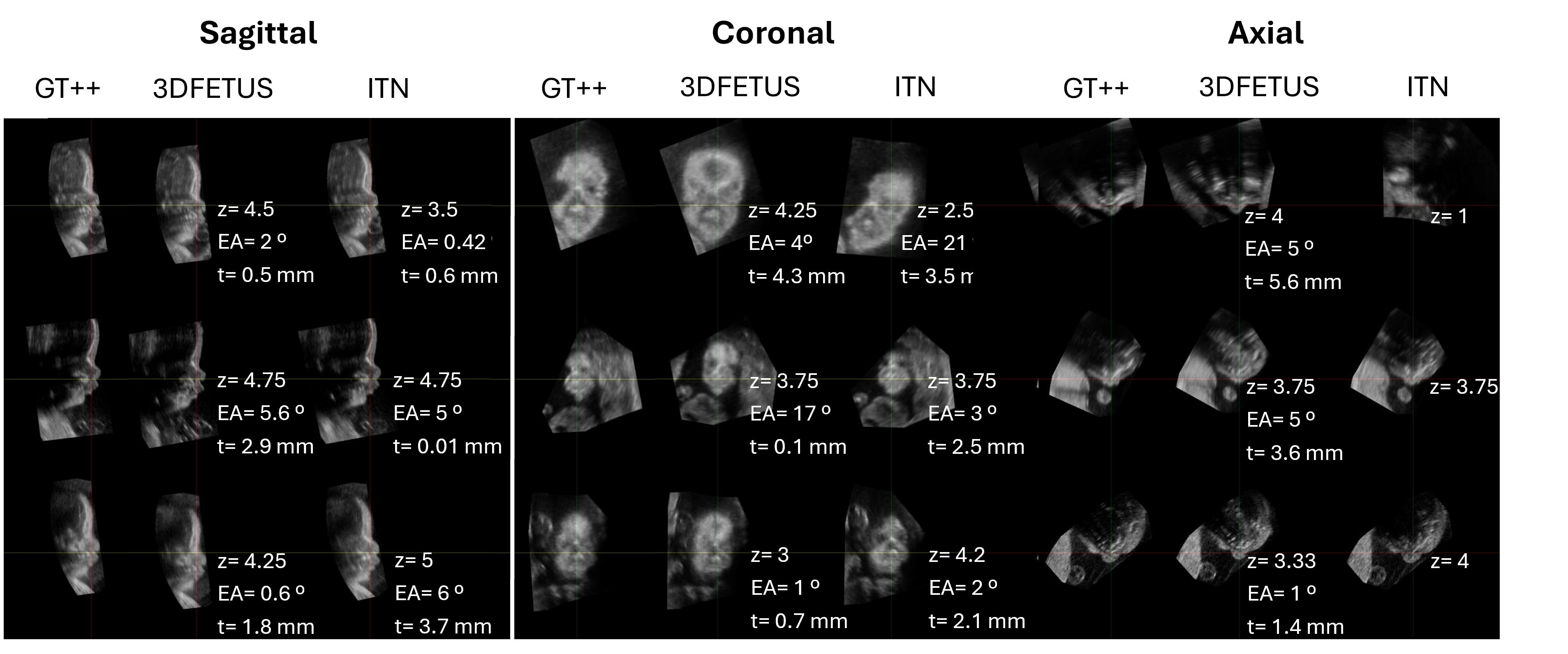}
\caption{Boxplots of rotation (a) and translation (b) errors on the test set. Comparison of 3DFETUS, ITN  \citep{Li2018} , and acquisition error relative to GT facial planes.} \label{fig:boxplot_methods}
\end{figure}
We evaluated global rotation and translation errors in Table~\ref{tab:global_error}. For a more detailed analysis, per-plane errors are reported in Table~\ref{tb: plane_error}.
The proposed method outperforms the inter-observer variability (without landmarks completion), significantly reduces both rotation and translation errors relative to the acquisition baseline, and achieves superior performance compared to ITN \citep{Li2018}.
A visual representation of the results in Table~\ref{tab:global_error} is provided in Figure~\ref{fig:boxplot_methods}, which shows boxplots comparing the proposed method (3DFETUS) against ITN \citep{Li2018}, and the acquisition error. It can be observed that 3DFETUS consistently achieves lower rotation and translation errors, with reduced variance, indicating more stable and accurate standard plane estimation across samples. Note that the number of outliers is reduced considerably by the proposed method. Figure \ref{fig:sota_3dfetus_comp} shows some qualitative examples of the predicted planes using the 3DFETUS and the ITN \citep{Li2018} method, next to the GT++ planes. 
\begin{figure}
\centering
\includegraphics[width=0.85\textwidth]{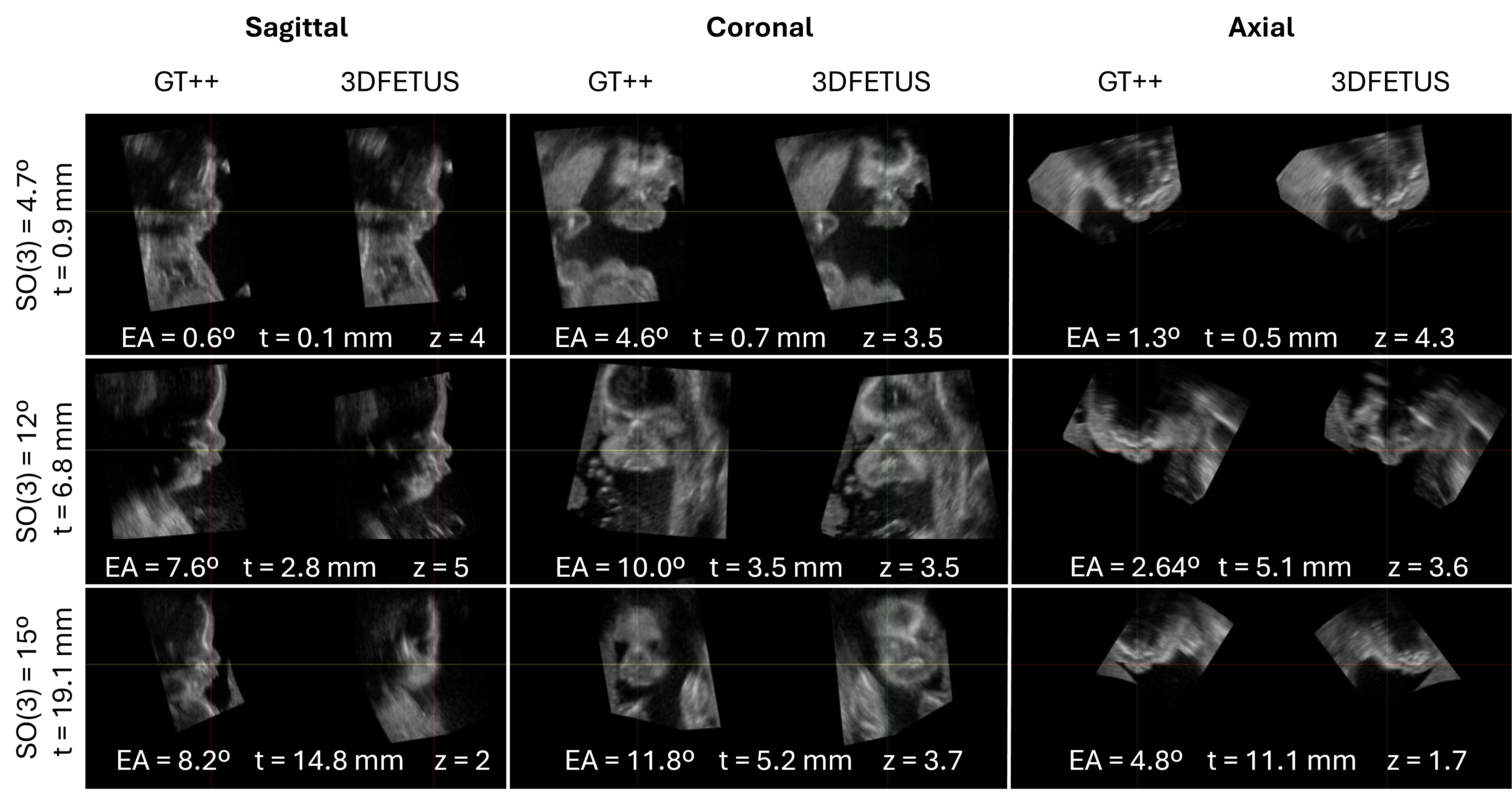}
\caption{3DFETUS performance on the test set. Displayed are 2D planes from three different test subjects representing high, average, and low performance of the proposed 3DFEUS method. For each subject, the Euclidean angle (EA, in degrees) and translation error ($t$, in mm) are reported with respect to the ground truth (GT). The mean quality score ($z$), as assigned by clinicians, is also indicated.} \label{fig:examples}
\end{figure}

\begin{figure}
\centering
\includegraphics[width=0.85\textwidth]{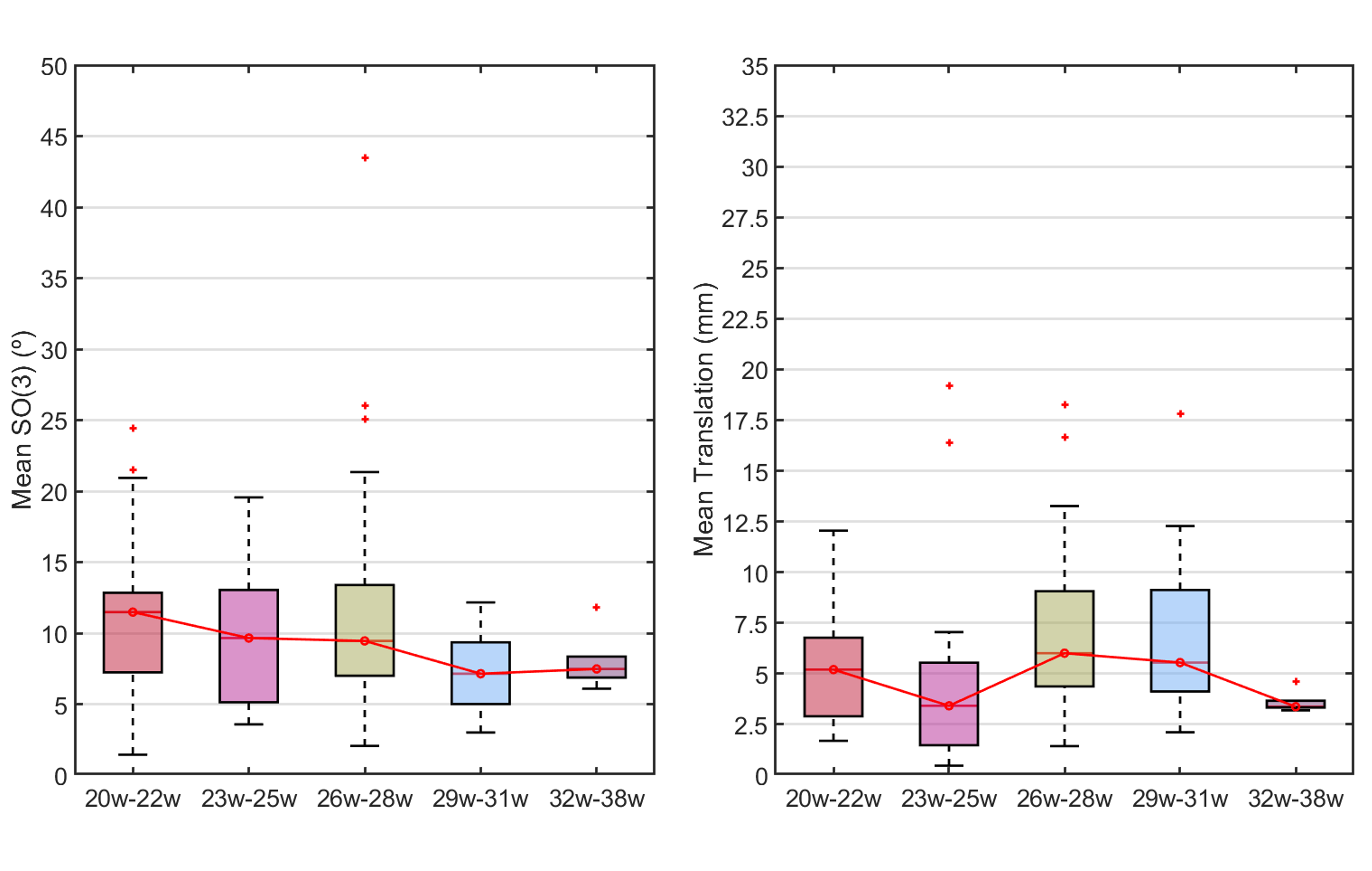}
\vspace{-3mm}
\caption{Boxplots of rotation (a) and translation (b) errors by gestational age. Results correspond to 3DFETUS predictions on the test set.} \label{fig:ga_analysis}
\end{figure}

\textbf{Gestational Age Analysis}:
Figure~\ref{fig:ga_analysis} presents boxplots illustrating the distribution of rotation (S0(3)) and translation errors across these gestational groups. The results show a slight decrease in rotation error as GA increases. Although the mean translation error exhibits slightly greater variability across GA groups, no clear overall trend is observed. One-way ANOVA analysis indicates that neither the rotation error ($p = 0.065$) nor the translation error ($p = 0.092$) shows statistically significant differences across GA groups. These findings suggest that the performance of the model is reasonably stable within the analyzed range of GAs.

\begin{table}
\centering
\caption{Quantitative test set comparison per plane. } \label{tb: plane_error} 
\begin{tabular}{|l|l|l|l|l|}
\hline
\textbf{} &\textbf{Model} & \textbf{Sagittal} & \textbf{Axial} &\textbf{Coronal}\\
\cline{2-5}
\multirow{3}{*}{$t$} & Acq & 14.85 
$\pm$ 7.11 & 9.02 $\pm$ 5.92 &  6.58 $\pm$ 4.98 \\
\cline{2-5}
& Inter-obs & 1.60 $\pm$ 1.74 & 1.94 $\pm$ 1.84 & 1.51 $\pm$ 1.40\\
\cline{2-5}
(mm)& ITN \citep{Li2018} & 7.16 $\pm$ 13.03 & 7.98$\pm$ 17.32 & 6.52 $\pm$ 12.28\\
\cline{2-5}
& 3DFETUS  & \textbf{3.28 $\pm$ 3.10} & \textbf{2.92 $\pm$ 2.46} & \textbf{4.11 $\pm$ 3.08}\\
\hline
\multirow{3}{*}{$R$} & Acq & 5.85 $\pm$ 6.26 & 62.09 $\pm$ 60.17 &  94.88 $\pm$ 13.60 \\
\cline{2-5}
& Inter-obs & 11.35 $\pm$ 21.66 &  7.08 $\pm$ 9.60 &  12.63 $\pm$ 26.73\\
\cline{2-5}
($^{\circ}$)& ITN \citep{Li2018} & 24.35 $\pm$ 46.21& 13.63 $\pm$ 20.89 & 25.63 $\pm$ 44.49\\
\cline{2-5}
& 3DFETUS  &\textbf{5.90 $\pm$ 4.48} & \textbf{5.68 $\pm$ 4.65}& \textbf{4.61 $\pm$ 5.73}\\
\hline
\end{tabular}
\end{table}

\textbf{Clinical Quality Score}:
Table \ref{tb: score_3dfetus} summarizes the mean quality scores and the standard deviation of the proposed 3DFETUS and ITN \citep{Li2018}. Both models are trained using the same data split. The results show how the clinicians give significant higher scores to the planes estimated using our proposed methodology. Considering a quality score threshold of 3 or higher as acceptable, the proposed 3DFETUS method achieved good facial plane standarization results in 87\% of the test set, with the remaining 13\% classified as low quality and bad facial planes. In comparison, ITN \citep{Li2018} attained 71\% good-quality results and 29\% low quality. These findings indicate that the proposed method improves the proportion of acceptable quality scans by 16\% over the existing approach.
\begin{table}
\centering
\caption{Qualitative comparison results of  3DFETUS and ITN \citep{Li2018}. Indicated with an * (paired t-test $p < 0.01$).} \label{tb: score_3dfetus} 
\begin{tabular}{|l|l|l|l|l|}
\hline
\textbf{Model} &\textbf{Global} & \textbf{Sagittal} & \textbf{Coronal} &\textbf{Axial}\\
\hline
3DFETUS & 4.05$\pm$0.73$^*$ & 4.04$\pm$0.73$^*$ & 3.57$\pm$0.68$^*$ & 3.42$\pm$0.74$^*$\\
\hline
ITN \citep{Li2018} & 3.51 $\pm$ 1.37 & 3.51$\pm$ 1.38 & 3.30$\pm$ 1.18 & 3.02 $\pm$ 1.16\\
\hline
\end{tabular}
\end{table}
\begin{table}
\centering
\caption{Loss Ablation study.} \label{tb: ablation_loss} 
\begin{tabular}{|l|l|l|l|l|}
\hline
\textbf{Metric} &\textbf{Model} & \textbf{Sagittal} & \textbf{Axial} &\textbf{Coronal}\\
\hline
$t$ &$\mathcal{L}_{\alpha r + \beta t}$ & 7.26 $\pm$ 6.57 & 5.78 $\pm$ 4.20 & 5.76 $\pm$ 4.78\\
\cline{2-5}
(mm) & $\mathcal{L}_{Grid}$ & \textbf{3.28 $\pm$ 3.10} & \textbf{2.92 $\pm$ 2.46} & \textbf{4.11 $\pm$ 3.08}\\
\hline
$R$ & $\mathcal{L}_{\alpha r + \beta t}$ & 8.70 $\pm$ 19.69& 6.14 $\pm$ 6.73& 7.49 $\pm$ 18.88\\
\cline{2-5}
($^{\circ}$)& $\mathcal{L}_{Grid}$ & \textbf{5.90 $\pm$ 4.48} & \textbf{5.68 $\pm$ 4.65}& \textbf{4.61 $\pm$ 5.73}\\
\hline
\end{tabular}
\end{table}

\textbf{Ablation Loss}:
 The results in Table \ref{tb: ablation_loss} indicate that incorporating the grid loss improves both translation and rotation estimation errors. The proposed grid-based loss helps balance the contributions of rotation and translation during training, providing good overall performance without the need for extensive hyperparameter tuning or manually finding a trade-off between the two components.

\subsection{Quantitative-Qualitative relation}
\begin{figure}
\centering
\includegraphics[width=0.65\textwidth]{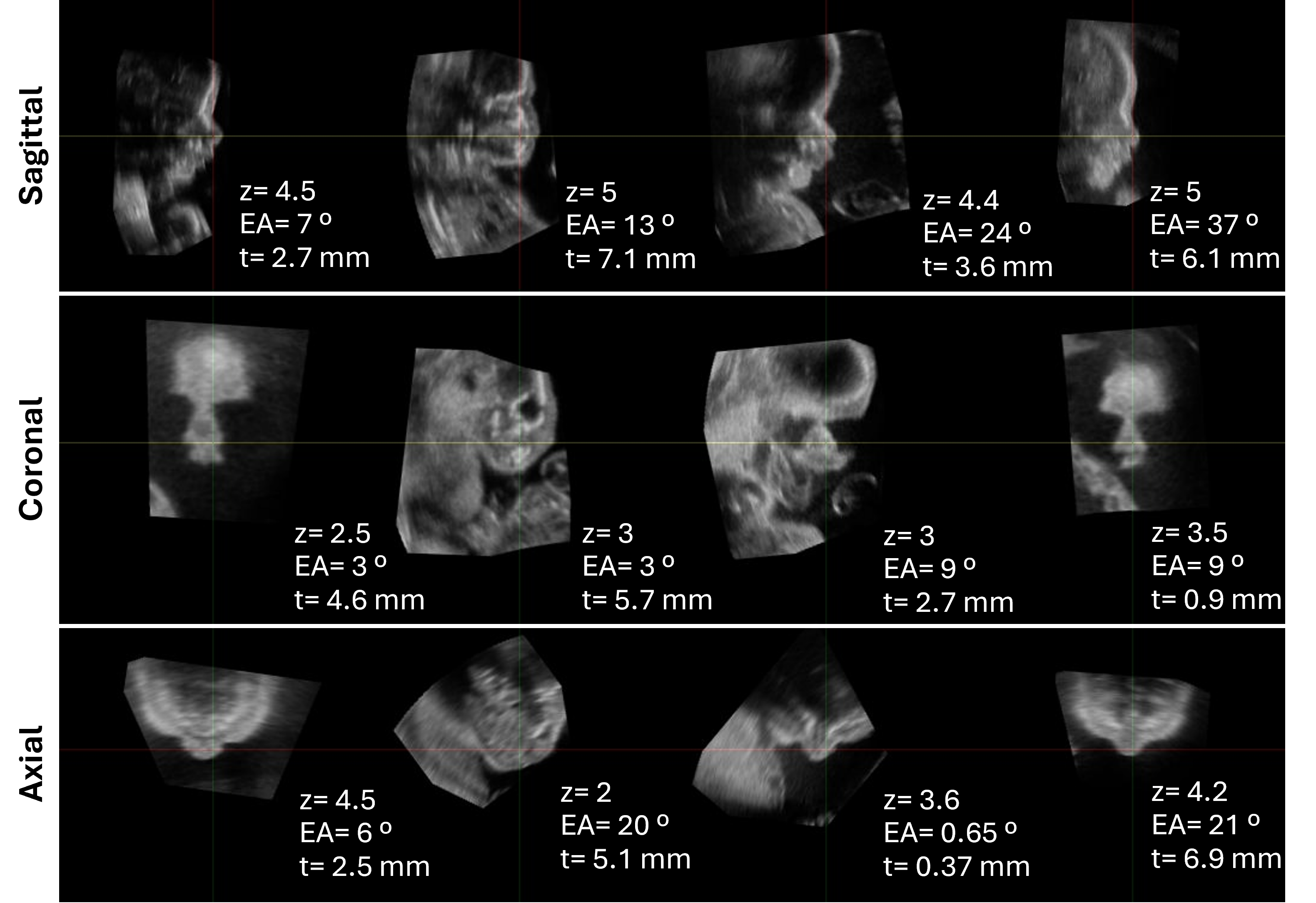}
\caption{Analysis examples of qualitative vs. quantitative facial plane quality. $z$ representing clinician-assigned mean quality scores, $t$ and EA corresponding Euclidean translation and rotation errors. 
}  \label{fig:ex_ql_vs_ql}
\end{figure}

\begin{figure}
\centering
\includegraphics[width=0.65\textwidth]{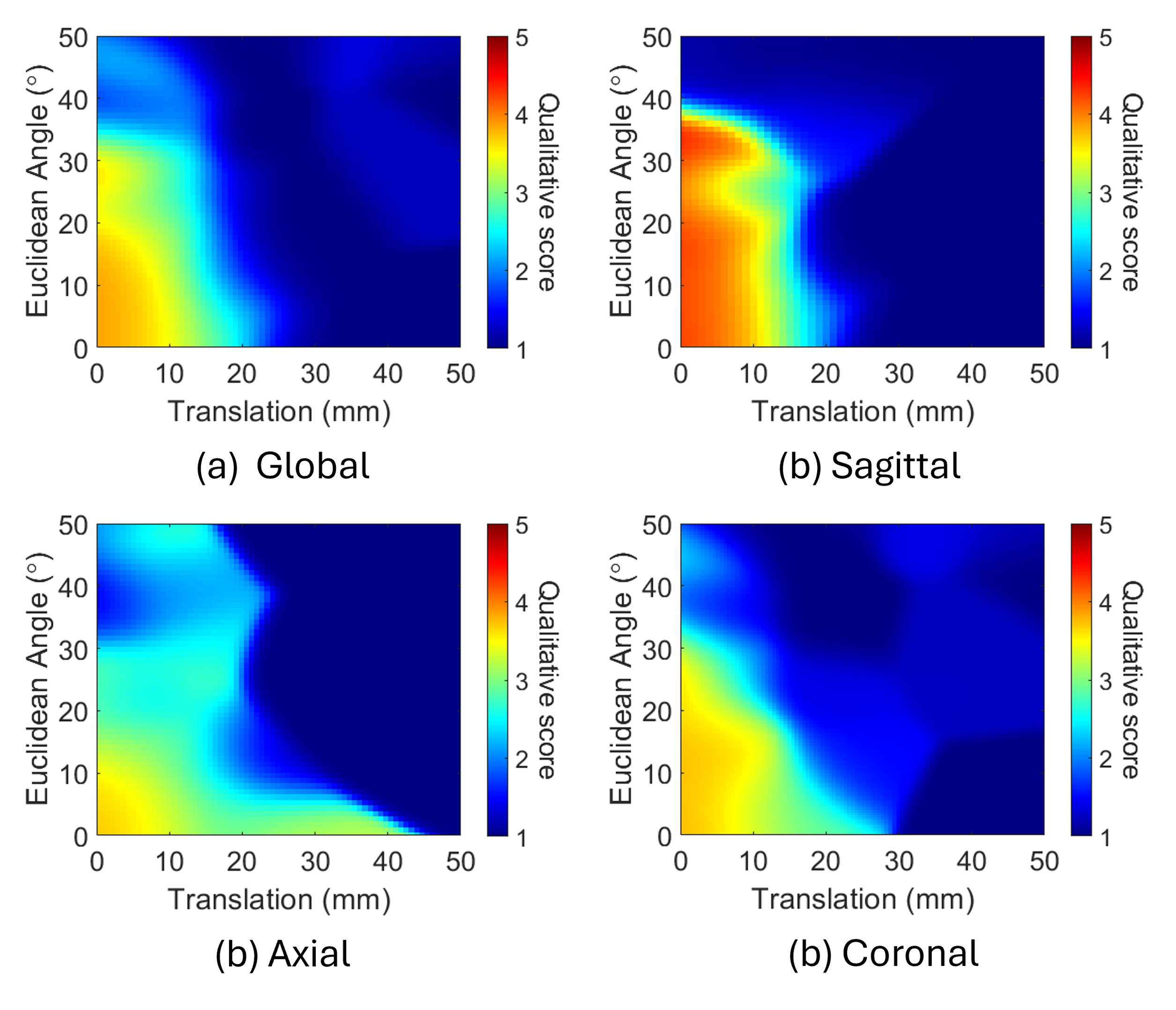}
\vspace{-3mm}
\caption{Comparison between qualitative and quantitative results across standard planes. The plots display translation error and rotation error on the axes; color indicates the mean quality score assigned by clinicians for scans with those errors. 
} \label{fig:qt_vs_ql}
\end{figure}

Figure \ref{fig:qt_vs_ql} illustrates the relationship between quality scores and quantitative errors in translation and rotation angle for facial SPs. The plot delineates the boundaries within which the facial SPs are considered clinically acceptable. Specifically, Figure \ref{fig:qt_vs_ql} presents this relationship aggregating all planes together, showing that regions with translation errors below 10 mm and rotation errors under 20$^{\circ}$ correspond to quality scores near 4. Beyond these error thresholds, quality scores progressively decline, indicating suboptimal facial SPs no longer deemed clinically acceptable.

Examining each plane individually reveals additional insights. In the sagittal plane plot, quality scores are generally high, consistent with the results in Table \ref{tb: score_3dfetus}. Notably, the quality scores are more sensitive to translation errors than to rotation errors. Planes with rotation errors as high as 30–40 degrees relative to the ground truth still receive high quality scores. This suggests that rotational discrepancies are difficult to discern in sagittal views, likely because anatomical structures vary less along this plane. To further illustrate this observation, Figure \ref{fig:ex_ql_vs_ql} presents several cases with varying quality scores alongside their corresponding rotation and translation errors. It becomes evident that some rotational errors are less perceptible in sagittal images, while other planes reveal them more clearly. For instance, even with significant rotation error in the respective facial plane, the sagittal image may remain visually similar to a high-quality, low-error example.

In contrast, axial planes receive more critical assessments from clinicians, with generally lower quality scores. These scores are sensitive to both translation and rotation errors, indicating that axial views provide stronger visual cues for detecting rotational misalignments and significant anatomical changes caused by translation.

Finally, coronal plane evaluations suggest intermediate sensitivity to translation and rotation errors, higher than the sagittal plane but lower than the axial plane. Quality scores in the coronal plane are better than those in the axial and align more closely with the overall global scores. This likely reflects the moderate impact of rotation and translation on anatomical structures visible in this view.

\section{Discussion}
The proposed GT++ algorithm to estimate the facial planes from landmark annotations is robust and has been clinically evaluated. By leveraging a 3DMM of newborns, GT++ improves facial plane estimation, particularly through the completion of missing landmarks. Clinicians’ qualitative evaluations confirmed this improvement, with the mean score increasing from 3.50 to 4.06 in scale from 1 to 5. The inter-observer study further demonstrates that this approach substantially reduces rotation errors, both in mean and standard deviation, stabilizing the results and minimizing outliers. 

The resulting DL model 3DFETUS demonstrates improved computational efficiency and precision compared to other state-of-the-art methods. It significantly reduces both rotation and translation errors while also achieving higher clinical evaluation scores.

The network successfully standardizes 3D US planes, obtaining a good qualitative score in 87\% of the cases. Evaluation across GA shows consistent performance, with no statistically significant differences observed, indicating that the model extracts facial features that are robust the developmental stage within the analyzed range (20-36 GA weeks). Given the inherent noise in US imaging, this suggests the model does not overfit to features specific to a particular GA. Although some variability was observed in translation performance, this was not found statistically significant and may be attributed to the effect of noise in the GT plane centers, which can introduce slight inconsistencies in translation estimation.

The qualitative versus quantitative analysis suggests that the sagittal view alone is insufficient for reliable facial plane standardization. Clinicians often assigned high qualitative scores even when the rotation error was as high as 30$^{\circ}$. This, along with visual examples, highlights that the sagittal plane is less descriptive and less reliable for assessing facial pose, as facial features do not vary significantly across this view. In contrast, the coronal and axial views proved to be more informative for evaluating facial pose quality, showing a stronger correlation with quantitative errors. From this analysis, we also observed that clinicians generally consider a facial plane to be well-standardized when the rotation error is below 20$^{\circ}$ and the translation error is under 10 mm. However, the axial view showed comparatively lower performance, likely because even slight translations can shift the anatomy substantially, producing a noticeably different slice where the anatomy differs. 

The mean performance of our method falls within this clinically accepted range, supporting its validity and reliability for automatic facial plane standardization in 3D US. Thus, the proposed network correctly learns the plane angles but its translation error is higher than the inter-observer error. This could be due to the high variability defining the planes localization, as multiple slices could closely resemble each other \citep{Skelton2021}. However, it could also be that the most informative planes for the network are not the ones defined as GT planes and it sacrifices translation accuracy for rotation accuracy. Although the translation errors obtained with the proposed network are slightly larger than the inter-observer variability, the mean translation error per plane is 3.21 $\pm$ 1.98mm. Importantly, rotation is generally more critical than translation, as the aim is to standardize the facial axes. By correcting for fetal pose and probe orientation, the network enables sonographers to examine a standardized 3D fetal facial US volume, facilitating more consistent and accurate assessment of the fetal face structures. Furthermore, the proposed network is able to reduce the rotation inter-observer variability, obtaining a mean euclidean error of 5.31 $\pm$ 3.945$^\circ$ per plane. This can be highly beneficial to homogenize the facial analysis evaluation criteria and to reduce the reliance on clinician expertise while reducing the time burden of manually locating the planes.  

While our study demonstrates the pipeline on fetal US, the approach provides a structured way to standardize volumetric data. In principle, similar strategies could be explored for other imaging tasks or organs where reliable plane localization is important, although further validation would be required.

\section{Conclusions}
We presented GT++ and 3DFETUS, two complementary components of a unified framework for a robust and automated standardization of 3D fetal facial US volumes. GT++ provides a robust method for estimating facial SPs from landmark annotations. By inferring the missing landmarks using a landmarks-based 3DMM of infants, GT++ significantly improves facial SPs quality, reduces inter-observer variability, and enhances clinical consistency. 3DFETUS builds upon GT++ to automatize the standardization of the 3D fetal US axes to the canonical front-right facial pose using DL. It outperforms state-of-the-art methods in both precision and computational efficiency, showing strong robustness to GA and effectively reducing rotation and translation errors. Together, GT++ and 3DFETUS provide a clinically validated and efficient pipeline for automating 3D US fetal face pose standardization, facilitating more consistent and prenatal facial evaluation.
%
%
%





\section*{Data Availability}
The datasets of the study are derived from clinical 3D fetal ultrasound volumes and cannot be publicly shared due to patient privacy and ethical restrictions.

\section*{Code Availability}
The code implementing the proposed algorithms, will be made publicly available upon publication to facilitate reproducibility and further research.

\section*{Acknowledgment}
This work is partly supported by MICIU/AEI/10.13039/501100011033/ under the project grant PID2024-161292OB-I00 and PRE2021-097544 scholarship, the ICREA Academia programme.



\bibliographystyle{elsarticle-harv} 
\bibliography{bib_complete}

@article{Gower1975,
   author = {J. C. Gower},
   doi = {10.1007/BF02291478},
   issn = {00333123},
   issue = {1},
   journal = {Psychometrika},
   title = {Generalized {Procrustes} analysis},
   volume = {40},
   year = {1975}
}

@article{Salomon2022,
   author = {L. J. Salomon and Z. Alfirevic and V. Berghella and C. M. Bilardo and G. E. Chalouhi and F. Da Silva Costa and E. Hernandez-Andrade and G. Malinger and H. Munoz and D. Paladini and F. Prefumo and A. Sotiriadis and A. Toi and W. Lee},
   doi = {10.1002/uog.24888},
   issn = {14690705},
   issue = {6},
   journal = {Ultrasound in Obstetrics and Gynecology},
   title = {ISUOG Practice Guidelines (updated): performance of the routine mid-trimester fetal ultrasound scan},
   volume = {59},
   year = {2022}
}

@article{Salomon2011,
   author = {L. J. Salomon and Z. Alfirevic and V. Berghella and C. Bilardo and E. Hernandez-Andrade and S. L. Johnsen and K. Kalache and K. Y. Leung and G. Malinger and H. Munoz and F. Prefumo and A. Toi and W. Lee},
   doi = {10.1002/uog.8831},
   issn = {09607692},
   issue = {1},
   journal = {Ultrasound in Obstetrics and Gynecology},
   title = {Practice guidelines for performance of the routine mid-trimester fetal ultrasound scan},
   volume = {37},
   year = {2011}
}

@article{Merz2017,
   author = {Eberhard Merz and Sonila Pashaj},
   doi = {10.1515/jpm-2016-0379},
   issn = {1619-3997},
   issue = {6},
   journal = {Journal of Perinatal Medicine},
   month = {1},
   title = {Advantages of {3D} ultrasound in the assessment of fetal abnormalities},
   volume = {45},
   year = {2017},
}

@article{Dyson2000,
   author = {R. L. Dyson and D. H. Pretorius and N. E. Budorick and D. D. Johnson and M. S. Sklansky and C. J. Cantrell and S. Lai and T. R. Nelson},
   doi = {10.1046/j.1469-0705.2000.00183.x},
   issn = {0960-7692},
   issue = {4},
   journal = {Ultrasound in Obstetrics \& Gynecology},
   month = {9},
   pages = {321-328},
   title = {Three‐dimensional ultrasound in the evaluation of fetal anomalies},
   volume = {16},
   year = {2000},
}

@article{Chen2022,
   author = {Jing Chen and Sangam Kanekar},
   doi = {10.1016/j.clp.2022.04.005},
   issn = {00955108},
   issue = {3},
   journal = {Clinics in Perinatology},
   month = {9},
   pages = {771-790},
   title = {Imaging of Congenital Craniofacial Anomalies and Syndromes},
   volume = {49},
   year = {2022},
}

@article{Bartzela2017,
   author = {Theodosia N. Bartzela and Carine Carels and Jaap C. Maltha},
   doi = {10.3389/fphys.2017.01038},
   issn = {1664-042X},
   journal = {Frontiers in Physiology},
   month = {12},
   title = {Update on 13 Syndromes Affecting Craniofacial and Dental Structures},
   volume = {8},
   year = {2017},
}

@article{Junaid2022,
   author = {Mohammed Junaid and Linda Slack-Smith and Kingsley Wong and Jenny Bourke and Gareth Baynam and Hanny Calache and Helen Leonard},
   doi = {10.1038/s41390-022-02024-9},
   issn = {0031-3998},
   issue = {6},
   journal = {Pediatric Research},
   month = {12},
   pages = {1795-1804},
   title = {{Association between craniofacial anomalies, intellectual disability and autism spectrum disorder: Western Australian population-based study}},
   volume = {92},
   year = {2022},
}

@inproceedings{Jaderberg_2015,
author = {Jaderberg, Max and Simonyan, Karen and Zisserman, Andrew and Kavukcuoglu, Koray},
title = {Spatial transformer networks},
year = {2015},
publisher = {MIT Press},
address = {Cambridge, MA, USA},
booktitle = {Proceedings of the 28th International Conference on Neural Information Processing Systems - Volume 2},
pages = {2017–2025},
numpages = {9},
location = {Montreal, Canada},
series = {NIPS'15}
}

@article{Sarris2011,
   author = {I. Sarris and C. Ioannou and M. Dighe and A. Mitidieri and M. Oberto and W. Qingqing and J. Shah and S. Sohoni and W. Al Zidjali and L. Hoch and D. G. Altman and A. T. Papageorghiou},
   doi = {10.1002/uog.8997},
   issn = {0960-7692},
   issue = {6},
   journal = {Ultrasound in Obstetrics \& Gynecology},
   month = {12},
   pages = {681-687},
   title = {Standardization of fetal ultrasound biometry measurements: improving the quality and consistency of measurements},
   volume = {38},
   year = {2011},
}

@ARTICLE{Chen_2017,
  author={Chen, Hao and Wu, Lingyun and Dou, Qi and Qin, Jing and Li, Shengli and Cheng, Jie-Zhi and Ni, Dong and Heng, Pheng-Ann},
  journal={IEEE Transactions on Cybernetics}, 
  title={Ultrasound Standard Plane Detection Using a Composite Neural Network Framework}, 
  year={2017},
  volume={47},
  number={6},
  pages={1576-1586},
  doi={10.1109/TCYB.2017.2685080}}

@article{Baumgartner2017,
   author = {Christian F. Baumgartner and Konstantinos Kamnitsas and Jacqueline Matthew and Tara P. Fletcher and Sandra Smith and Lisa M. Koch and Bernhard Kainz and Daniel Rueckert},
   doi = {10.1109/TMI.2017.2712367},
   issn = {0278-0062},
   issue = {11},
   journal = {IEEE Transactions on Medical Imaging},
   month = {11},
   pages = {2204-2215},
   title = {{SonoNet: Real-Time Detection and Localisation of Fetal Standard Scan Planes in Freehand Ultrasound}},
   volume = {36},
   year = {2017},
}

@inbook{Li2018_2,
   author = {Yuanwei Li and Bishesh Khanal and Benjamin Hou and Amir Alansary and Juan J. Cerrolaza and Matthew Sinclair and Jacqueline Matthew and Chandni Gupta and Caroline Knight and Bernhard Kainz and Daniel Rueckert},
   doi = {10.1007/978-3-030-00928-1\_45},
   pages = {392-400},
   title = {Standard Plane Detection in {3D} Fetal Ultrasound Using an Iterative Transformation Network},
   year = {2018},
}

@article{Li2018,
   author = {Y Li and J Cerrolaza and M Sinclair and B Hou and A Alansary and B Khanal and J Matthew and B Kainz and D Rueckert},
   issue = {Midl},
   journal = {Medical Imaging with Deep Learning (MIDP)},
   title = {{Standard Plane Localisation in 3D Fetal Ultrasound Using Network with Geometric and Image Loss}},
   year = {2018}
}

@article{Sarris2012,
   author = {I. Sarris and C. Ioannou and P. Chamberlain and E. Ohuma and F. Roseman and L. Hoch and D. G. Altman and A. T. Papageorghiou},
   doi = {10.1002/uog.10082},
   issn = {0960-7692},
   issue = {3},
   journal = {Ultrasound in Obstetrics \& Gynecology},
   month = {3},
   pages = {266-273},
   title = {Intra‐ and interobserver variability in fetal ultrasound measurements},
   volume = {39},
   year = {2012},
}

@inproceedings{Li2021,
   author = {Keyu Li and Jian Wang and Yangxin Xu and Hao Qin and Dongsheng Liu and Li Liu and Max Q.-H. Meng},
   doi = {10.1109/ICRA48506.2021.9561295},
   isbn = {978-1-7281-9077-8},
   journal = {2021 IEEE International Conference on Robotics and Automation (ICRA)},
   month = {5},
   pages = {8302-8308},
   publisher = {IEEE International Conference on Robotics and Automation},
   title = {Autonomous Navigation of an Ultrasound Probe Towards Standard Scan Planes with Deep Reinforcement Learning},
   year = {2021},
}

@article{Skelton2021,
   author = {E. Skelton and J. Matthew and Y. Li and B. Khanal and J.J. Cerrolaza Martinez and N. Toussaint and C. Gupta and C. Knight and B. Kainz and J.V. Hajnal and M. Rutherford},
   doi = {10.1016/j.radi.2020.11.006},
   issn = {10788174},
   issue = {2},
   journal = {Radiography},
   month = {5},
   pages = {519-526},
   title = {Towards automated extraction of {2D}  standard fetal head planes from {3D} ultrasound acquisitions: A clinical evaluation and quality assessment comparison},
   volume = {27},
   year = {2021},
}

@article{Lei2015,
   author = {Baiying Lei and Ee-Leng Tan and Siping Chen and Liu Zhuo and Shengli Li and Dong Ni and Tianfu Wang},
   doi = {10.1371/journal.pone.0121838},
   issn = {1932-6203},
   issue = {5},
   journal = {PLOS ONE},
   month = {5},
   pages = {e0121838},
   title = {Automatic Recognition of Fetal Facial Standard Plane in Ultrasound Image via {Fisher Vector}},
   volume = {10},
   year = {2015},
}

@inproceedings{Feng_2009,
   author = {Shaolei Feng and S. Kevin Zhou and Sara Good and Dorin Comaniciu},
   doi = {10.1109/CVPR.2009.5206527},
   isbn = {978-1-4244-3992-8},
   journal = {2009 IEEE Conference on Computer Vision and Pattern Recognition},
   month = {6},
   pages = {2488-2495},
   publisher = {IEEE},
   title = {Automatic fetal face detection from ultrasound volumes via learning {3D} and {2D}  information},
   year = {2009},
}

@article{Nerea_2024,
   author = {Nerea González-Aranceta and Antonia Alomar and Ricardo Rubio and Silvia Maya-Enero and Antonio Payá and Gemma Piella and Federico Sukno},
   doi = {10.1016/j.earlhumdev.2024.106021},
   issn = {03783782},
   journal = {Early Human Development},
   month = {6},
   pages = {106021},
   title = {Accuracy and repeatability of fetal facial measurements in {3D} ultrasound: A longitudinal study},
   volume = {193},
   year = {2024},
}

@article{Schlemper_2019,
title = {Attention gated networks: Learning to leverage salient regions in medical images},
journal = {Medical Image Analysis},
volume = {53},
pages = {197-207},
year = {2019},
issn = {1361-8415},
doi = {https://doi.org/10.1016/j.media.2019.01.012},
author = {Jo Schlemper and Ozan Oktay and Michiel Schaap and Mattias Heinrich and Bernhard Kainz and Ben Glocker and Daniel Rueckert}
}

@article{Di_Vece_2024,
  author={DiVece, Chiara and Lous, Maela Le and Dromey, Brian and Vasconcelos, Francisco and David, Anna L. and Peebles, Donald and Stoyanov, Danail},
  journal={IEEE Transactions on Medical Robotics and Bionics}, 
  title={Ultrasound Plane Pose Regression: Assessing Generalized Pose Coordinates in the Fetal Brain}, 
  year={2024},
  volume={6},
  number={1},
  pages={41-52},
  doi={10.1109/TMRB.2023.3328638}}

@article{Zhen2023,
   author = {Chaojiong Zhen and Hongzhang Wang and Jun Cheng and Xin Yang and Chaoyu Chen and Xindi Hu and Yuanji Zhang and Yan Cao and Dong Ni and Weijun Huang and Ping Wang},
   doi = {10.1016/j.ultrasmedbio.2023.05.005},
   issn = {03015629},
   issue = {9},
   journal = {Ultrasound in Medicine \& Biology},
   month = {9},
   pages = {2006-2016},
   title = {Locating Multiple Standard Planes in First-Trimester Ultrasound Videos via the Detection and Scoring of Key Anatomical Structures},
   volume = {49},
   year = {2023},
}

@InProceedings{He2021,
   author = {Shuangchi He and Zehui Lin and Xin Yang and Chaoyu Chen and Jian Wang and Xue Shuang and Ziwei Deng and Qin Liu and Yan Cao and Xiduo Lu and Ruobing Huang and Nishant Ravikumar and Alejandro Frangi and Yuanji Zhang and Yi Xiong and Dong Ni},
   doi = {10.1007/978-3-030-87589-3\_20},
   pages = {190-198},
   title = {Statistical Dependency Guided Contrastive Learning for Multiple Labeling in Prenatal Ultrasound},
    booktitle="Machine Learning in Medical Imaging",
   year = {2021},
}

@inbook{Zou2022,
   author = {Yuxin Zou and Haoran Dou and Yuhao Huang and Xin Yang and Jikuan Qian and Chaojiong Zhen and Xiaodan Ji and Nishant Ravikumar and Guoqiang Chen and Weijun Huang and Alejandro F. Frangi and Dong Ni},
   doi = {10.1007/978-3-031-16440-8\_29},
   pages = {300-309},
   title = {Agent with Tangent-Based Formulation and Anatomical Perception for Standard Plane Localization in {3D} Ultrasound},
   year = {2022},
}

@inbook{Huang2020,
   author = {Yuhao Huang and Xin Yang and Rui Li and Jikuan Qian and Xiaoqiong Huang and Wenlong Shi and Haoran Dou and Chaoyu Chen and Yuanji Zhang and Huanjia Luo and Alejandro Frangi and Yi Xiong and Dong Ni},
   doi = {10.1007/978-3-030-59716-0\_53},
   pages = {553-562},
   title = {Searching Collaborative Agents for Multi-plane Localization in {3D} Ultrasound},
   year = {2020},
}

@inbook{Dou2019,
   author = {Haoran Dou and Xin Yang and Jikuan Qian and Wufeng Xue and Hao Qin and Xu Wang and Lequan Yu and Shujun Wang and Yi Xiong and Pheng-Ann Heng and Dong Ni},
   doi = {10.1007/978-3-030-32254-0\_33},
   pages = {290-298},
   title = {Agent with Warm Start and Active Termination for Plane Localization in {3D} Ultrasound},
   year = {2019},
}

@article{Benacerraf2019,
   author = {Beryl R. Benacerraf and Bryann Bromley and Angie C. Jelin},
   doi = {10.1016/j.ajog.2019.08.048},
   issn = {10976868},
   issue = {5},
   journal = {American Journal of Obstetrics and Gynecology},
   title = {{SMFM Fetal Anomalies Consult Series 1: Facial Anomalies}},
   volume = {221},
   year = {2019}
}

@article{Lamanna2023,
   author = {Bruno Lamanna and Miriam Dellino and Eliano Cascardi and Mia Rooke-Ley and Marina Vinciguerra and Gerardo Cazzato and Antonio Malvasi and Amerigo Vitagliano and Pierpaolo Nicolì and Michele Di Cosola and Andrea Ballini and Ettore Cicinelli and Antonella Vimercati},
   doi = {10.3390/jcm12165365},
   issn = {20770383},
   issue = {16},
   journal = {Journal of Clinical Medicine},
   title = {Efficacy of Systematic Early-Second-Trimester Ultrasound Screening for Facial Anomalies: A Comparison between Prenatal Ultrasound and Postmortem Findings},
   volume = {12},
   year = {2023}
}

@article{Mak2019,
   author = {Annisa Shui Lam Mak and Kwok Yin Leung},
   doi = {10.14366/usg.18031},
   issn = {22885943},
   issue = {1},
   journal = {Ultrasonography},
   title = {Prenatal ultrasonography of craniofacial abnormalities},
   volume = {38},
   year = {2019}
}

@article{Chen2024,
   author = {Zeyuan Chen and Yuanjie Zheng and James C. Gee},
   doi = {10.1109/TMI.2023.3288136},
   issn = {1558254X},
   issue = {1},
   journal = {IEEE Transactions on Medical Imaging},
   title = {TransMatch: A Transformer-Based Multilevel Dual-Stream Feature Matching Network for Unsupervised Deformable Image Registration},
   volume = {43},
   year = {2024}
}

@article{DEVOS2019128,
title = {A deep learning framework for unsupervised affine and deformable image registration},
journal = {Medical Image Analysis},
volume = {52},
pages = {128-143},
year = {2019},
issn = {1361-8415},
doi = {https://doi.org/10.1016/j.media.2018.11.010},
url = {https://www.sciencedirect.com/science/article/pii/S1361841518300495},
author = {Bob D. {de Vos} and Floris F. Berendsen and Max A. Viergever and Hessam Sokooti and Marius Staring and Ivana Išgum}
}

@article{Balakrishnan2019,
   author = {Guha Balakrishnan and Amy Zhao and Mert R. Sabuncu and John Guttag and Adrian V. Dalca},
   doi = {10.1109/TMI.2019.2897538},
   issn = {1558254X},
   issue = {8},
   journal = {IEEE Transactions on Medical Imaging},
   title = {VoxelMorph: A Learning Framework for Deformable Medical Image Registration},
   volume = {38},
   year = {2019}
}

@article{DiVece2022,
   author = {Chiara DiVece and Brian Dromey and Francisco Vasconcelos and Anna L. David and Donald Peebles and Danail Stoyanov},
   doi = {10.1007/s11548-022-02609-z},
   issn = {1861-6429},
   issue = {5},
   journal = {International Journal of Computer Assisted Radiology and Surgery},
   month = {5},
   pages = {833-839},
   title = {Deep learning-based plane pose regression in obstetric ultrasound},
   volume = {17},
   year = {2022}
}

@article{DiVece2024,
   author = {Chiara DiVece and Maela Le Lous and Brian Dromey and Francisco Vasconcelos and Anna L. David and Donald Peebles and Danail Stoyanov},
   doi = {10.1109/TMRB.2023.3328638},
   issn = {2576-3202},
   issue = {1},
   journal = {IEEE Transactions on Medical Robotics and Bionics},
   month = {2},
   pages = {41-52},
   title = {Ultrasound Plane Pose Regression: Assessing Generalized Pose Coordinates in the Fetal Brain},
   volume = {6},
   year = {2024}
}

@article{Dou2025,
   author = {Haoran Dou and Yuhao Huang and Yunzhi Huang and Xin Yang and Chaojiong Zhen and Yuanji Zhang and Yi Xiong and Weijun Huang and Dong Ni},
   doi = {10.1016/j.cmpb.2025.108619},
   issn = {01692607},
   journal = {Computer Methods and Programs in Biomedicine},
   month = {4},
   pages = {108619},
   title = {Standard plane localization using denoising diffusion model with multi-scale guidance},
   volume = {261},
   year = {2025}
}

@article{Namburete2018,
   author = {Ana I.L. Namburete and Weidi Xie and Mohammad Yaqub and Andrew Zisserman and J. Alison Noble},
   doi = {10.1016/j.media.2018.02.006},
   issn = {13618415},
   journal = {Medical Image Analysis},
   month = {5},
   pages = {1-14},
   title = {Fully-automated alignment of {3D}  fetal brain ultrasound to a canonical reference space using multi-task learning},
   volume = {46},
   year = {2018}
}

@inproceedings{Alvarez-Tun2023,
   author = {Olaya Álvarez-Tuñón and Yury Brodskiy and Erdal Kayacan},
   booktitle = {Europe ISR 2023 - International Symposium on Robotics, Proceedings},
   title = {{Loss it right: Euclidean and Riemannian Metrics in Learning-based Visual Odometry}},
   year = {2023}
}

@inproceedings{Zhou2019,
   author = {Yi Zhou and Connelly Barnes and Jingwan Lu and Jimei Yang and Hao Li},
   doi = {10.1109/CVPR.2019.00589},
   issn = {10636919},
   booktitle = {Proceedings of the IEEE Computer Society Conference on Computer Vision and Pattern Recognition},
   title = {On the continuity of rotation representations in neural networks},
   volume = {2019-June},
   year = {2019}
}

@inproceedings{Park2022,
   author = {Jaewoo Park and Nam Ik Cho},
   doi = {10.1007/978-3-031-20068-7\_21},
   issn = {16113349},
   booktitle = {Lecture Notes in Computer Science (including subseries Lecture Notes in Artificial Intelligence and Lecture Notes in Bioinformatics)},
   title = {DProST: Dynamic Projective Spatial Transformer Network for 6D Pose Estimation},
   volume = {13666 LNCS},
   year = {2022}
}

@article{Men2025,
   author = {Qianhui Men and He Zhao and Lior Drukker and Aris T. Papageorghiou and J. Alison Noble},
   doi = {10.1016/j.media.2025.103614},
   issn = {13618415},
   journal = {Medical Image Analysis},
   month = {8},
   pages = {103614},
   title = {ScanAhead: Simplifying standard plane acquisition of fetal head ultrasound},
   volume = {104},
   year = {2025}
}

@article{Espinoza2013,
   author = {Jimmy Espinoza and Sara Good and Evie Russell and Wesley Lee},
   doi = {10.7863/ultra.32.5.847},
   issn = {15509613},
   issue = {5},
   journal = {Journal of Ultrasound in Medicine},
   title = {Does the use of automated fetal biometry improve clinical work flow efficiency?},
   volume = {32},
   year = {2013}
}

@article{Liu2019,
   author = {Shengfeng Liu and Yi Wang and Xin Yang and Baiying Lei and Li Liu and Shawn Xiang Li and Dong Ni and Tianfu Wang},
   doi = {10.1016/j.eng.2018.11.020},
   issn = {20958099},
   issue = {2},
   journal = {Engineering},
   title = {Deep Learning in Medical Ultrasound Analysis: A Review},
   volume = {5},
   year = {2019}
}

@article{Matthew2022,
   author = {Jacqueline Matthew and Emily Skelton and Thomas G. Day and Veronika A. Zimmer and Alberto Gomez and Gavin Wheeler and Nicolas Toussaint and Tianrui Liu and Samuel Budd and Karen Lloyd and Robert Wright and Shujie Deng and Nooshin Ghavami and Matthew Sinclair and Qingjie Meng and Bernhard Kainz and Julia A. Schnabel and Daniel Rueckert and Reza Razavi and John Simpson and Jo Hajnal},
   doi = {10.1002/pd.6059},
   issn = {10970223},
   issue = {1},
   journal = {Prenatal Diagnosis},
   title = {Exploring a new paradigm for the fetal anomaly ultrasound scan: Artificial intelligence in real time},
   volume = {42},
   year = {2022}
}

@inproceedings{Singh2021,
   author = {Tejal Singh and Srinivas Rao Kudavelly and K. Venkata Suryanarayana},
   doi = {10.1109/ISBI48211.2021.9433915},
   issn = {19458452},
   booktitle = {Proceedings - International Symposium on Biomedical Imaging},
   title = {Deep learning based fetal face detection and visualization in prenatal ultrasound},
   volume = {2021-April},
   year = {2021}
}

@inproceedings{Yu2016,
   author = {Zhen Yu and Dong Ni and Siping Chen and Shengli Li and Tianfu Wang and Baiying Lei},
   doi = {10.1109/EMBC.2016.7590780},
   issn = {1557170X},
   booktitle = {Proceedings of the Annual International Conference of the IEEE Engineering in Medicine and Biology Society, EMBS},
   title = {Fetal facial standard plane recognition via very deep convolutional networks},
   volume = {2016-October},
   year = {2016}
}

@inproceedings{Chen2020,
   author = {Chaoyu Chen and Xin Yang and Ruobing Huang and Wenlong Shi and Shengfeng Liu and Mingrong Lin and Yuhao Huang and Yong Yang and Yuanji Zhang and Huanjia Luo and Yankai Huang and Yi Xiong and Dong Ni},
   doi = {10.1109/ISBI45749.2020.9098368},
   issn = {19458452},
   booktitle = {Proceedings - International Symposium on Biomedical Imaging},
   title = {Region Proposal Network with Graph Prior and Iou-Balance Loss for Landmark Detection in 3D Ultrasound},
   volume = {2020-April},
   year = {2020}
}

@article{RamirezZegarra2023,
   author = {R. Ramirez Zegarra and T. Ghi},
   doi = {10.1002/uog.26130},
   issn = {14690705},
   issue = {2},
   journal = {Ultrasound in Obstetrics and Gynecology},
   title = {Use of artificial intelligence and deep learning in fetal ultrasound imaging},
   volume = {62},
   year = {2023}
}

@article{Kwitt2013,
   author = {R. Kwitt and N. Vasconcelos and S. Razzaque and S. Aylward},
   doi = {10.1016/j.media.2013.05.003},
   issn = {13618415},
   issue = {7},
   journal = {Medical Image Analysis},
   title = {Localizing target structures in ultrasound video - A phantom study},
   volume = {17},
   year = {2013}
}

@article{Maraci2014,
   author = {Mohammad Ali Maraci and Raffaele Napolitano and Aris Papageorghiou and J. Alison Noble},
   doi = {10.1007/978-3-319-10581-9\_17},
   issn = {16113349},
   journal = {Lecture Notes in Computer Science },
   title = {Searching for structures of interest in an ultrasound video sequence},
   volume = {8679},
   year = {2014}
}

@inproceedings{Rahmatullah2012,
   author = {Bahbibi Rahmatullah and Aris T. Papageorghiou and J. Alison Noble},
   doi = {10.1007/978-3-642-33454-2\_50},
   issn = {16113349},
   booktitle = {Lecture Notes in Computer Science},
   title = {Integration of local and global features for anatomical object detection in ultrasound},
   volume = {7512},
   year = {2012}
}

@article{Ni2014,
   author = {Dong Ni and Xin Yang and Xin Chen and Chien Ting Chin and Siping Chen and Pheng Ann Heng and Shengli Li and Jing Qin and Tianfu Wang},
   doi = {10.1016/j.ultrasmedbio.2014.06.006},
   issn = {1879291X},
   issue = {11},
   journal = {Ultrasound in Medicine and Biology},
   title = {Standard Plane Localization in Ultrasound by Radial Component Model and Selective Search},
   volume = {40},
   year = {2014}
}

@article{Rathbun2021,
   author = {Kimberly M. Rathbun and Aaron S. Zweig},
   doi = {10.1016/j.jemermed.2021.07.048},
   issn = {07364679},
   issue = {6},
   journal = {The Journal of Emergency Medicine},
   month = {12},
   pages = {744-748},
   title = {A Realistic and Inexpensive Ultrasound Phantom for Teaching M-Mode Measurement of Fetal Heart Rate},
   volume = {61},
   year = {2021}
}

@article{Duan2025,
   author = {Yaofei Duan and Tao Tan and Zhiyuan Zhu and Yuhao Huang and Yuanji Zhang and Rui Gao and Patrick Cheong-Iao Pang and Xinru Gao and Guowei Tao and Xiang Cong and Zhou Li and Lianying Liang and Guangzhi He and Linliang Yin and Xuedong Deng and Xin Yang and Dong Ni},
   doi = {10.1016/j.media.2025.103725},
   issn = {13618415},
   journal = {Medical Image Analysis},
   month = {10},
   pages = {103725},
   title = {FetalFlex: Anatomy-guided diffusion model for flexible control on fetal ultrasound image synthesis},
   volume = {105},
   year = {2025}
}

@article{Qiao2026,
   author = {Sibo Qiao and Mengru Huang and Gang Luo and Wenjing Yin and Hengxiao Li and Min Wang and Zhihan Lyu},
   doi = {10.1016/j.eswa.2025.128939},
   issn = {09574174},
   journal = {Expert Systems with Applications},
   month = {1},
   pages = {128939},
   title = {Fetal ultrasound four-chamber view editing synthesis via denoising diffusion model},
   volume = {295},
   year = {2026}
}

@article{Otsu1979,
   author = {Nobuyuki Otsu},
   doi = {10.1109/TSMC.1979.4310076},
   issn = {0018-9472},
   issue = {1},
   journal = {IEEE Transactions on Systems, Man, and Cybernetics},
   month = {1},
   pages = {62-66},
   title = {A Threshold Selection Method from Gray-Level Histograms},
   volume = {9},
   year = {1979}
}

@article{Morales2025,
   author = {Araceli Morales and Antonia Alomar and Antonio R. Porras and Marius George Linguraru and Gemma Piella and Federico M. Sukno},
   doi = {{https://doi.org/10.1016/j.compbiomed.2025.109652}},
   issn = {00104825},
   journal = {Computers in Biology and Medicine},
   month = {3},
   pages = {109652},
   title = {{BabyFM: Towards accurate 3D baby facial models using spectral decomposition and asymmetry swapping}},
   volume = {186},
   year = {2025},
}

@article{WADDINGTON2025,
title = {3D imaging and geometric morphometrics of facial dysmorphology and asymmetry indicate gestational timings of dysmorphogenesis in schizophrenia and bipolar disorder},
journal = {European Neuropsychopharmacology},
volume = {93},
pages = {1-2},
year = {2025},
issn = {0924-977X},
doi = {https://doi.org/10.1016/j.euroneuro.2024.12.007},
author = {John L. Waddington and Federico M. Sukno}
}
\end{document}